\documentclass[lettersize,journal]{IEEEtran}
\usepackage{amsmath,amsfonts}
\usepackage{algorithmic}
\usepackage{algorithm}
\usepackage{array}
\usepackage[caption=false,font=normalsize,labelfont=sf,textfont=sf]{subfig}
\usepackage{textcomp}
\usepackage{stfloats}
\usepackage{url}
\usepackage{verbatim}
\usepackage{graphicx}
\usepackage{cite}
\usepackage{booktabs}    % 用于 \toprule, \midrule 和 \bottomrule
\usepackage{tabularx}    % 用于 tabularx 环境
\usepackage{multirow}  % 用于合并单元格
\usepackage{xcolor}
\usepackage{caption}
\usepackage{bibentry}
\usepackage{stfloats}  % 引入 stfloats 包来控制浮动体
\usepackage{amssymb}  % 引入 amssymb 包
\usepackage[table]{xcolor}
\usepackage{pifont}
\usepackage{makecell}
\begin{document}

\title{OmniScaleSR: Unleashing Scale-Controlled Diffusion Prior for Faithful and Realistic Arbitrary-Scale Image Super-Resolution}
\author{Xinning Chai, Zhengxue Cheng~\IEEEmembership{Member,~IEEE,}, Yuhong Zhang, Hengsheng Zhang, Yingsheng Qin, Yucai Yang, Rong Xie, ~\IEEEmembership{Member,~IEEE,} Li Song, ~\IEEEmembership{Senior Member,~IEEE}
        % <-this % stops a space
% \thanks{}% <-this % stops a space
\thanks{Xinning Chai, Zhengxue Cheng, Yuhong Zhang, and Rong Xie are with the School of Information Science and Electronic Engineering, Shanghai Jiao Tong University (E-mail: chaixinning@sjtu.edu.cn, zxcheng@sjtu.edu.cn, rainbowow@sjtu.edu.cn, xierong@sjtu.edu.cn).

Hengsheng Zhang is with the College of Information, Mechanical and Electrical Engineering, Shanghai Normal University (E-mail: zhang\_hs@shnu.edu.cn).

Yucai Yang and Yingsheng Qin are with Transsion in China.

Li Song is with the School of Information Science and Electronic Engineering, Shanghai Jiao Tong University and the MoE Key Lab of Artificial Intelligence, AI Institute, Shanghai Jiao Tong University, China (E-mail: song\_li@sjtu.edu.cn). \textit{(corresponding author: Zhengxue Cheng, Li Song)}
}}

% The paper headers
\markboth{Journal of \LaTeX\ Class Files,~Vol.~14, No.~8, August~2021}%
{Shell \MakeLowercase{\textit{et al.}}: A Sample Article Using IEEEtran.cls for IEEE Journals}

\IEEEpubid{0000--0000/00\$00.00~\copyright~2021 IEEE}
% Remember, if you use this you must call \IEEEpubidadjcol in the second
% column for its text to clear the IEEEpubid mark.

\maketitle

\begin{abstract}
Arbitrary-scale super-resolution (ASSR) overcomes the limitation of traditional super-resolution (SR) that works only at a fixed scale (e.g., ×4), enabling a single model to achieve arbitrary-scale SR. 
Most ASSR methods explicitly incorporate implicit neural representation (INR) to achieve ASSR, but INR's inherently regression-driven feature extraction and aggregation nature restricts their capacity to synthesize meticulous details, leading to low realism. 
Recently, diffusion-based realistic image super-resolution (Real-ISR) methods leverage the pre-trained diffusion prior and have shown promising results at ×4 scale. We find that they could also achieve ASSR because the powerful pre-trained diffusion prior implicitly employs SR scale adaptation by encouraging the model to always generate high-realism images. However, due to the lack of explicit SR scale controls, the model fails to effectively manage the diffusion behavior according to different SR scales, causing either excessive hallucination or blurry results, especially for ultra-high magnification.
To address these limitations, we propose \emph{OmniScaleSR}, a novel diffusion-based realistic
arbitrary-scale super-resolution (Real-ASSR) method to achieve both high fidelity and high-realism ASSR. We introduce explicit diffusion-native SR scale controls, which could be elegantly coupled with the implicit scale adaptation, unleashing scale-controlled diffusion prior to dynamically managing the diffusion behavior in a content- and scale-aware manner. 
Furthermore, we incorporate multi-domain fidelity enhancement designs to achieve more faithful reconstruction. Extensive experiments on both bicubic degradation benchmarks and real-world datasets demonstrate that OmniScaleSR consistently outperforms state-of-the-art methods in terms of both fidelity and perceptual realism, with especially strong performance under high-magnification scenarios. Codes will be at https://github.com/chaixinning/OmniScaleSR.

\end{abstract}


\section{Introduction}
\IEEEpubidadjcol
\IEEEPARstart{I}{mage} Super-Resolution (SR) aims to reconstruct a high-resolution (HR) image from its low-resolution (LR) version. Traditional image SR methods are restricted to fixed SR scales. Arbitrary-Scale Super-Resolution (ASSR) seeks to overcome it and enables a single unified model to handle arbitrary SR scales. LIIF \cite{chen2021learning} is a pioneering work that introduces implicit neural representations (INR) to achieve ASSR. It predicts the RGB value at an arbitrary query coordinate by taking an image coordinate and local features around the coordinate. Building on this, followed ASSR methods \cite{lee2022local,chen2023cascaded,cao2023ciaosr,song2023ope,he2024latent} modify the architecture, feature aggregation, and position encoding to improve the performance. GanssianSR \cite{hu2025gaussiansr} overcomes the inherent discontinuity
of pixel-based INR methods by representing each pixel as a continuous Gaussian field. However, all of these methods use a pretrained regression-optimized CNN/Transformer encoder to extract features, usually EDSR \cite{lim2017enhanced}, RDN \cite{zhang2018residual} or SwinIR \cite{liang2021swinir}, which exhibits limited capacity to model complex distributions with fine high-frequency details, causing blurry results with low realism.

\IEEEpubidadjcol
Diffusion models \cite{sohl2015deep,ho2020denoising} have demonstrated a powerful capability to model complex data distributions. IDM \cite{gao2023implicit} integrates implicit neural representations (INR) into a pixel-domain diffusion model and designs a dual denoising branch to complement the local feature extraction of EDSR. IND \cite{kim2024arbitrary} improves efficiency by employing a latent diffusion model, relocating the INR component from the repetitive denoising process to the VAE decoder. However, both methods still rely on the INR-based scale control paradigm originally designed for CNN- and Transformer-based ASSR frameworks, leaving the intrinsic regression-based feature extraction and aggregation mechanisms unchanged. Consequently, they fail to fully leverage the diffusion's generative ability, leading to significantly degraded perceptual quality at large SR scales, and even apparent blocking artifacts appear.

\begin{figure*}[t] % [h] 表示图片将尽可能放置在当前位置（here）
% \begin{subfigure}
    \centering % 图片居中
    \includegraphics[width=\textwidth]{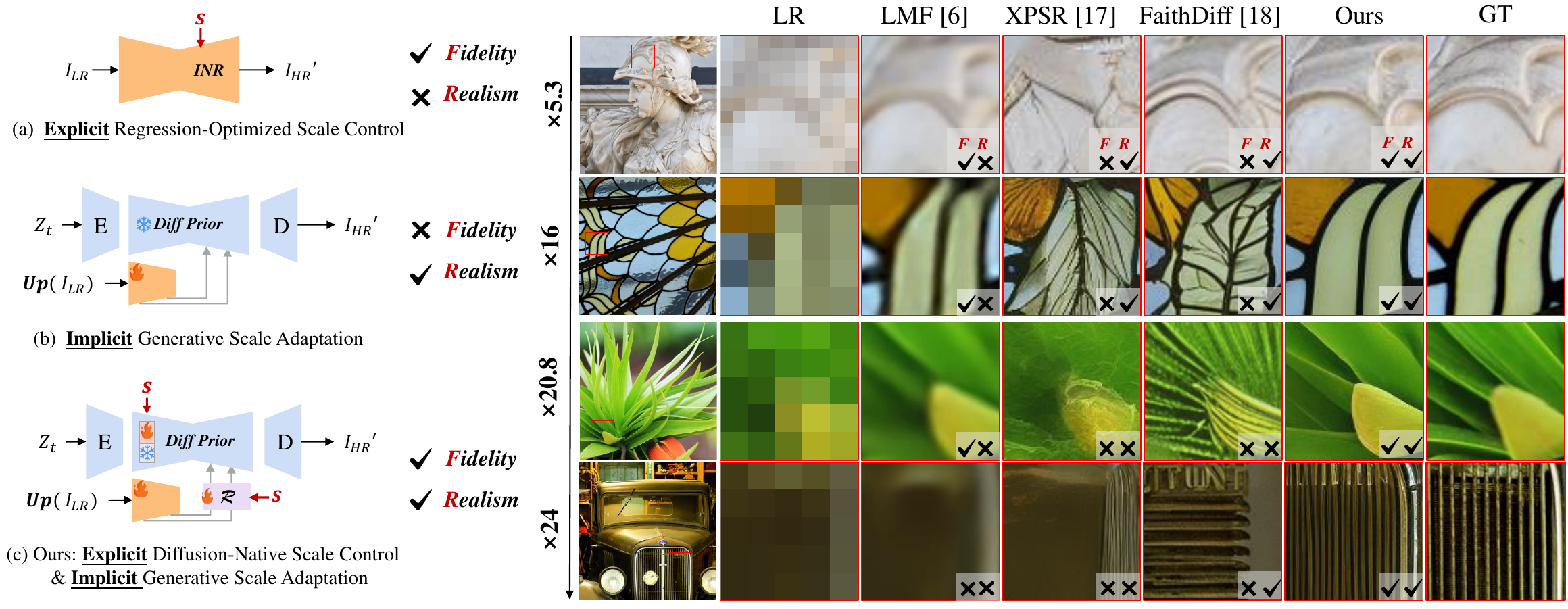} % 调整图片宽度为页面宽度的80%
    \caption{\textbf{Left:} Comparison of ASSR implementations in existing methods: (a) Explicit INR-based scale control with regression-based optimization \cite{chen2021learning,lee2022local,chen2023cascaded,cao2023ciaosr,song2023ope,he2024latent,gao2023implicit,kim2024arbitrary} (b) Implicit generative scale adaptation benefitted from pre-trained diffusion prior in diffusion-based Real-ISR methods \cite{lin2024diffbir, yang2025pixel, yu2024scaling, wu2024seesr,sun2024coser,qu2024xpsr,chen2025faithdiff}, (c) Our OmniScaleSR adopts both explicit diffusion-native scale controls and implicit generative scale adaptation for both high-fidelity and high-realism ASSR. \textbf{Right:} Visual comparison with the state-of-the-art methods. 'F' and 'R' are abbreviations of 'Fidelity' and 'Realism'.} % 添加图片标题
    \label{fig:teaser} % 添加图片标签，便于引用
    % \vspace{-0.5em}
% \end{subfigure}
\end{figure*}

Recently, generative diffusion models \cite{rombach2022high} have demonstrated impressive
generative capabilities in image synthesis and image editing, which inspires recent diffusion-based realistic image super-resolution (Real-ISR) methods (e.g., SeeSR \cite{wu2024seesr}, XPSR \cite{qu2024xpsr}, FaithDiff \cite{chen2025faithdiff}) to leverage the pre-trained Text-to-Image (T2I) prior for high-realism SR with meticulous and realistic details. They could also achieve ASSR by directly upscaling LR images to target resolutions and then enhancing them. We find that the powerful pre-trained diffusion prior implicitly achieves SR scale adaptation by encouraging the model always to generate high-realism images. Nevertheless, relying solely on implicit scale adaptation fails to fundamentally manage the generative dynamics and is prone to over- or under-activating its generative ability, leading to \textit{inconsistent semantics with chaotic details} or \textit{blurry results}, as shown in Fig. \ref{fig:teaser}, resulting in low fidelity.

We posit that the expected diffusion behavior exhibits substantial variation across different SR scales. 
At large SR scales, we expect the model to sufficiently activate generative priors to recover massive structural and textural details since the LR input undergoes severe information loss. In contrast, at small SR scales, the focus of the model shifts to suppress excessive hallucinations and improve fidelity since most image contents can be directly learned from the LR image.
Therefore, the explicit SR scale control is urgent to manage the diffusion behavior across a wide range of SR scales and heterogeneous image contents.

\IEEEpubidadjcol
Based on the above analysis, we propose \textbf{OmniScaleSR}, a novel diffusion-based Realistic Arbitrary-Scale Super-Resolution (Real-ASSR) method to achieve both high fidelity and high realism ASSR simultaneously. The core idea of our method is to elegantly combine the explicit diffusion-native scale controls and implicit generative scale adaptation, thereby unleashing scale-controlled diffusion prior to dynamically adjusting the model's generation and fidelity capabilities based on the target SR scale and image contents, strengthening the generation capability at large SR scale and high-frequency details, and enhancing the fidelity capability at low SR scale and coarse structure.

To achieve this, we propose two explicit diffusion-native scale control mechanisms: a \textbf{global scale injection mechanism} to provide an overall perception, and a local scale modulation mechanism to dynamically control the coordination between the model's generation and fidelity capabilities in a content- and scale-aware manner. Unlike IND \cite{kim2024arbitrary}, which constrains the SR scale solely at the VAE decoder, the proposed global scale injection mechanism incorporates the SR scale awareness throughout the whole latent denoising process. For \textbf{local scale modulation mechanism}, as shown in Fig. \ref{fig:teaser} (c), we leverage a scale-modulated refinement network (denoted as $\mathcal{R}$) to coordinate our main generation branch and the fidelity branch according to the SR scale and spatial information. 
Furthermore, we employ \textbf{multi-domain fidelity enhancement designs} to further improve fidelity, including pixel-domain, pixel-to-latent domain, and the latent domain.

Our main contributions can be summarized as follows:

\begin{itemize}
  \item To the best of our knowledge, we are the first to unleash scale-controlled diffusion prior for Real-ASSR, elegantly integrating explicit diffusion-native scale controls with implicit generative scale adaptation for both high-fidelity and high-realism ASSR.
  
  \item We introduce two synergistic explicit scale control mechanisms to dynamically manage the latent diffusion behavior according to the SR scale and image contents: global scale injection mechanism and local scale modulation mechanism.

  \item We integrate multi-domain fidelity enhancement designs to stabilize high-fidelity controllable generation across both moderate and extreme SR scales, ensuring consistent image fidelity and realism.

  \item Extensive experiments across both bicubic-degraded benchmarks and real-world datasets demonstrate that our method surpasses all existing methods, especially at ultra-large SR scales.
\end{itemize}

\section{Related works}
\subsection{Single Image Super Resolution (SISR)}

Dong et al. \cite{dong2014learning} were the first to introduce deep learning into SISR, which inspired a surge of CNN-based approaches \cite{ledig2017photo, lim2017enhanced, zhang2018residual, zhang2018image, mei2021image}. 
SwinIR \cite{liang2021swinir} further applied Transformer architectures to address the limited receptive field problem inherent to CNNs. 
Building on this line of research, HAT \cite{chen2023activating} expanded the spatial modeling capability by introducing an overlapping cross-attention mechanism. 
To break the locality constraint of window-based attention, Zhan et al. \cite{zhang2024transcending} constructed a token dictionary to exploit long-range priors. 
More recently, Long et al. \cite{long2025progressive} proposed the Progressive Focused Transformer (PFT), which adaptively concentrates attention on the most informative tokens.

To address real-world data with complex degradations, BSRGAN \cite{zhang2021designing} introduces a randomized degradation shuffling strategy to increase the diversity of synthetic degradation patterns, while Real-ESRGAN \cite{wang2021realesrgan} adopts a high-order degradation modeling process to more closely align with real-world distributions. DASR \cite{liang2022efficient} presents a degradation-adaptive framework that dynamically modulates network parameters by estimating the input-specific degradation. FeMaSR \cite{chen2022real} enhances reconstruction quality by matching degraded LR image features with corresponding pre-trained HR priors. FSN \cite{guan2024frequency} proposes a frequency separation (FS) module based on Gaussian filtering to progressively decompose low-frequency components and synthesize high-frequency details. Recently, diffusion-based Real-ISR has gained impressive progress.

\subsection{Diffusion-based Super-Resolution}
Inspired by the non-equilibrium statistical physics, Sohl-Dickstein et al. \cite{sohl2015deep} firstly proposed the diffusion model to model complex distributions. The Denoising Diffusion Probabilistic Model (DDPM) \cite{ho2020denoising} later established a novel connection between diffusion models and denoising score matching, advancing the field significantly. Song et al. \cite{song2020score} proposed a unified framework for diffusion models, formulating them from the perspective of stochastic differential equations (SDEs). The Latent Diffusion Model (LDM) \cite{rombach2022high} extends the training of DDPMs to the latent space, enabling more efficient large-scale training for pre-trained text-to-image (T2I) diffusion models. These T2I models have shown immense potential not only in generating images but also in tasks such as image editing, video generation, and 3D content creation. Recently, diffusion priors have proven to be highly effective across various image restoration tasks as well \cite{luo2024controlling,zhang2025ssp,wei2024toward,welker2024driftrec, chan2024anlightendiff}.

SR3 \cite{saharia2022image} is a pioneering work that applied the denoising diffusion probabilistic models to single image SR. 
StableSR \cite{wang2024exploiting} leverages pre-trained text-to-image prior \cite{rombach2022high} as assistance for more realistic image SR, and proposes time-aware guidance to modulate the features by the timestep adaptively. 
DiffBIR \cite{lin2024diffbir} designs a two-stage method that first reconstructs a coarse image by a pre-trained SwinIR \cite{liang2021swinir} in the first stage, and then utilizes the diffusion model to refine textures and details in the second stage.
CoSeR \cite{sun2024coser} proposes to first create semantically coherent reference images as guidance to harness the implicit diffusion priors to enhance the LR input.
PASD \cite{yang2025pixel} proposes a pixel-aware cross-attention module to perceive pixel-level information, and it supports personalized stylization tasks like cartoonization by shifting the base model.
SUPIR \cite{yu2024scaling} explores the scaling law on the SR task. It empowers the model by model scaling, dataset enrichment, and multi-modal large language model assistance.
SeeSR \cite{wu2024seesr} proposes to utilize high-quality semantic prompts to enhance the generative capability of pre-trained T2I models for image SR and achieve more realistic detail generation. 
% SSP-IR \cite{zhang2025ssp} integrates the visual comprehension capabilities of MLLM and the visual representations from low-quality images to fully exploit semantic and structure priors. 
XPSR \cite{qu2024xpsr} utilizes Multimodal Large Language Models (MLLMs) LLaVA \cite{liu2023visual} to obtain cross-modal priors as guidance. Faithdiff \cite{chen2025faithdiff} utilizes SDXL \cite{podell2024sdxl} as the backbone and develops an alignment module to unleash the diffusion prior for faithful structures.

However, these methods lack explicit SR scale controls, failing to sufficiently manage the diffusion behavior for different SR scales, tending to synthesize either excessive hallucination or blurry results, especially at higher SR scales.

\begin{figure*}[h] % [h] 表示图片将尽可能放置在当前位置（here）
    \centering % 图片居中
    \includegraphics[width=\textwidth]{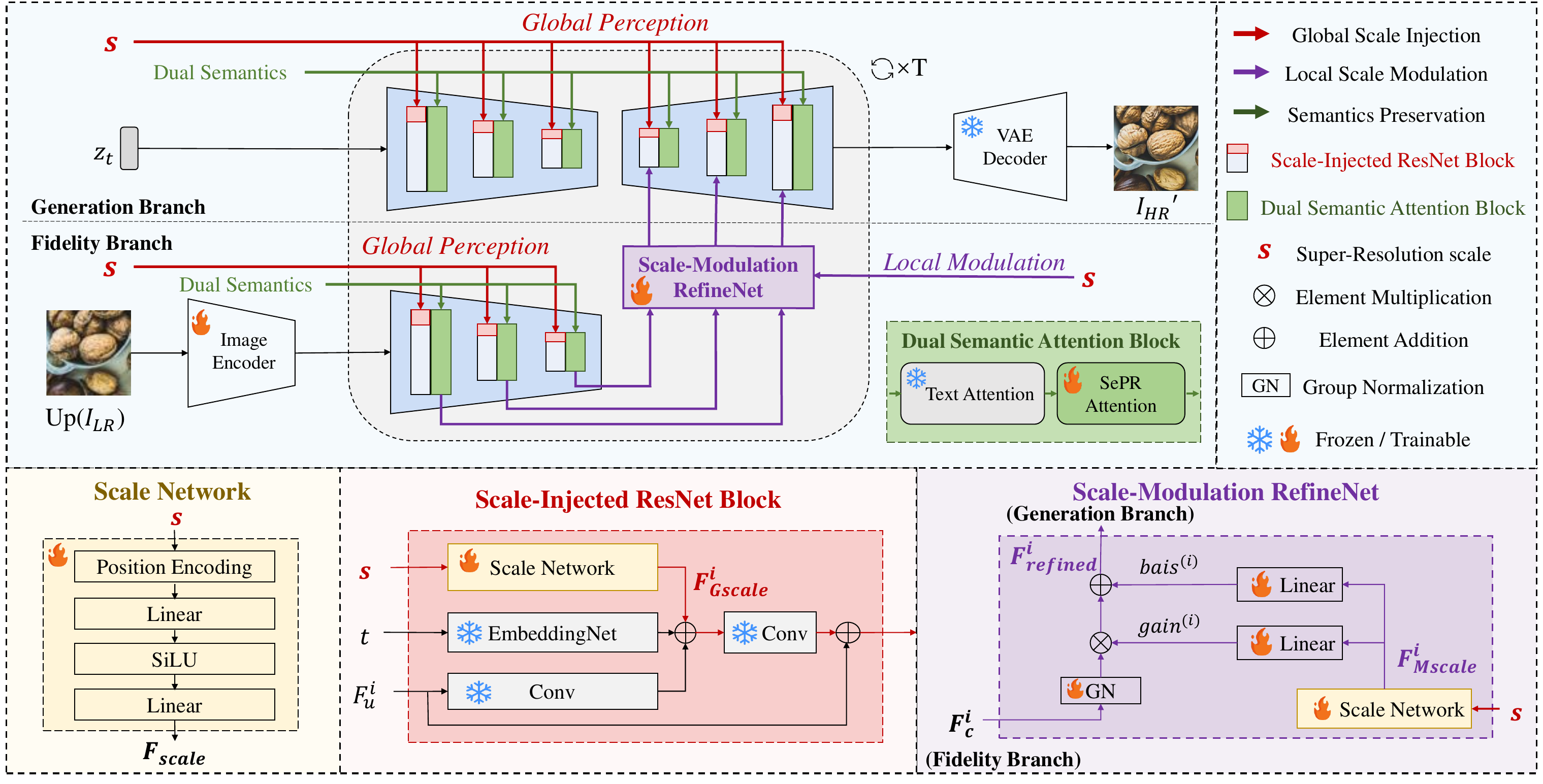} % 调整图片宽度为页面宽度的80%
    \caption{Overview of \textbf{OmniScaleSR}, which consists of a generation branch and a fidelity branch in the latent space. To enable explicit diffusion-native SR scale controls, we introduce a global scale injection mechanism (red arrow) for overall perception and a local scale modulation mechanism (purple arrow) to dynamically manage the model's generation and fidelity abilities.
    } % 添加图片标题
    \label{fig:overview} % 添加图片标签，便于引用
    \vspace{-1em}
\end{figure*}

\subsection{Arbitrary-Scale Super-Resolution (ASSR)}
Arbitrary-Scale Super-Resolution (ASSR) seeks to overcome the limitations of traditional SISR methods by enabling the model to handle arbitrary SR scales. It is first introduced in Meta-SR \cite{hu2019meta}, which designs a meta-upscale module that dynamically predicts the weights of upscale filters. LIIF \cite{chen2021learning} first employs implicit neural representations (INR) to predict RGB values at arbitrary query coordinates. Building on this,
Lee et al. \cite{lee2022local} introduce a local texture estimator to characterize image textures in 2D Fourier space. 
CLIT \cite{chen2023cascaded} integrates cross-scale local attention mechanism and frequency encoding technique into a local implicit Transformer. 
CiaoSR \cite{cao2023ciaosr} proposes a continuous implicit attention-in-attention network to ensemble nearby local features in a larger perception field. 
OPE-SR \cite{song2023ope} introduces orthogonal position encoding, and designs a parameter-free OPE-Upscale module for ASSR. 
LMF \cite{he2024latent} alleviates the INR computation cost by proposing a latent modulated function, which relocates the time-consuming operations and large MLPs to latent space. 
GaussianSR \cite{hu2025gaussiansr} represents each pixel as a continuous Gaussian field instead of discrete points and trains a classifier that adaptively assigns the appropriate 2D Gaussian kernel to each pixel.
Wei et al.\cite{wei2025multi} propose a scale-adaptive dynamic convolution and combine it with an implicit neural upsampling module to realize multi-contrast magnetic resonance imaging (MRI) super-resolution.
Extreme SR with ultra-high magnification remains one of the most challenging settings in ASSR. Wei et al. \cite{wei2024toward} is a pioneer work that recognizes the extreme-SR challenge and addresses it using a VQGAN prior, but it is limited to fixed-scale extreme SR.

Recent ASSR methods adopt the diffusion model to better model the complex distribution across the various SR scales. IDM \cite{gao2023implicit} integrates the INR into the denoising UNet decoder of the denoising diffusion model to achieve ASSR. 
IND \cite{kim2024arbitrary} adopts a latent diffusion architecture and relocates the implicit neural representation outside of the repeated diffusion process, combining it with the VAE decoder. E$^2$DiffSR \cite{wu2025latent} is an ASSR method for remote sensing images. It employs the VQ-VAE decoder features of the latent diffusion model to supplement the high-frequency details for the CNN-based ASSR branch, and proposes a feature modulation mechanism to adaptively balance contributions from the diffusion decoder branch and the CNN-based SR branch.

Although IDM, IND and E$^2$DiffSR incorporate explicit SR-scale conditioning into diffusion models, they still rely on INRs to realize ASSR, remaining essentially regression-based feature extraction and aggregation unchanged, failing to fully exploit the powerful diffusion's generative capability. Consequently, they struggle to recover fine details and often produce blurry results, especially at large SR scales. In this paper, we fully exploit the generative ability of the diffusion and introduce explicit scale control mechanisms to unleash scale-controlled diffusion prior for both high-fidelity and high-realism ASSR.

\section{Method}
\subsection{Overview of the framework}
Fig. \ref{fig:overview} illustrates the overall framework of OmniScaleSR. OmniScaleSR tackles ASSR challenges across varying and extreme SR scales by introducing explicit diffusion-native SR scale control mechanisms, enabling latent diffusion behavior to adapt to the target SR scale and image content. The proposed explicit scale control mechanisms organically cooperate and enhance the pre-trained diffusion prior, thereby enabling both high-fidelity and high-realism ASSR.

\begin{equation}
    I_{HR}' = \Phi(I_{LR}, s)
\end{equation}
where $\Phi$ represents the OmniScaleSR framework, and $s$ denotes the SR scale.

OmniScaleSR consists of two complementary branches in the latent space: a generative branch that generates the HR image from noise, and a fidelity branch that conditions the generation process on the LR image.

\subsection{Global Scale Injection Mechanism}
The target SR scale directly determines the amount of expected generation information. The global scale injection mechanism is designed to establish an overall perception for SR-scale-modulated generation. Specifically, we propagate the SR scale $s$ across each level of the multi-scale denoising features of both the generation branch and the fidelity branch.

To effectively perceive variations in continuous floating-point SR scales while avoiding excessive model complexity, we draw inspiration from the manner in which timestep information is incorporated in Stable Diffusion \cite{rombach2022high}. As shown in Fig. \ref{fig:overview}, we introduce a scale-injected ResNet block. In addition to the timestep information used in the original Stable Diffusion design, we further integrate SR scale information into the ResNet layers.
Specifically, we first convert the floating-point SR scale $s$ into a scale feature using a trainable scale network ${N}^i_{Gscale}$, then we add multi-scale scale features into the denoising UNet backbones:
\begin{equation}
    F_{Gscale}^i = \mathcal{N}_{Gscale}^i(s)
\end{equation}
\begin{equation}
    Fusion^i = Conv(Conv(F_u^i)\\+N^i_{emb}(t)+ \mathcal{N}^i_{Gscale}(s))
\end{equation}
where \( \mathcal{N}^i_{Gscale} \) denotes the scale network at level \( i \) of the denoising UNet, \( F_u^i \) represents the backbone features at level \( i \) in the denoising UNet, \( N^i_{emb} \) is the embedding network at level \( i \) to encode the timestep \( t \), and \( F^i_{Gscale} \) denotes the scale features at level \( i \) used for global perception.

The detailed architecture of the scale network is shown in Fig. \ref{fig:overview}. It is lightweight, consisting of two linear layers activated by the SiLU function:
\begin{equation}
    \mathcal{N}^i_{Gscale}(s) = Linear_1^i(SiLU(Linear_2^i(PE(s)))) \\
\end{equation}
where \( Linear \) denotes the linear layers and  \( PE \) refers to the sinusoidal position encoding.

We leverage the diffusion prior for scale adaptation by freezing the pre-trained denoising UNet and timestep embedding networks, training only the newly added scale networks. We apply the global scale injection mechanism for both the generation branch and the fidelity branch.

\subsection{Local Scale Modulation Mechanism}
The key to simultaneously achieving both high realism and high fidelity across different SR scales within a single model lies in a well-balanced coordination between the model’s generative and fidelity capacities. To regulate the diffusion behavior in a scale- and content-aware manner, we propose the local scale modulation mechanism.

As shown in Fig. \ref{fig:overview}, we insert a scale-modulated refinement network between the generation branch and the fidelity branch, called RefineNet. It serves as an explicit control module to dynamically adjust the cooperation between the model's fidelity and generation capabilities. 

Specifically, given the $i$-th layer feature $F_{c}^{i}$ in the fidelity branch, we first convert the SR scale $s$ into a feature $F^i_{Mscale}$ using a scale network $\mathcal{N}^i_{Mscale}$:
\begin{equation}
    F^i_{Mscale} = \mathcal{N}^i_{Mscale}(s) \\
\end{equation}

$\mathcal{N}^i_{Mscale}$ shares the same architecture as $\mathcal{N}^i_{Gscale}$ but different weights.
Then we learn modulation features from $F^i_{Mscale}$, including a gain modulation feature and a bias modulation feature:
\begin{equation}
    gain^{i} = Linear^{i}_{g}(F^i_{Mscale}),
\end{equation}
\begin{equation}
    bias^{i} = Linear^{i}_{b}(F^i_{Mscale}),
\end{equation}
where $Linear^{i}_{g}$ and $Linear^{i}_{b}$ represent linear layers.

Next we use them to modulate $F_{c}^{i}$:
\begin{equation}
    F^{i}_{refined} = gain^{i} \odot GroupNorm(F_{c}^{i}) + bias^{i}
\end{equation}
where $\odot$ denotes the channel-wise multiplication, and $GroupNorm$ denotes the group normalization. 

Finally, we add the multi-scale modulated spatial features $F^{i}_{refined}$ into the corresponding-level features in the main branch. All parameters in the scale-modulation RefineNet are trainable.

\subsection{Multi-Domain Fidelity Enhancement Designs}
As the SR scale increases, the challenge of preserving semantic and structural fidelity becomes even more difficult, particularly for ultra-high magnification. To address this, we strengthen the model fidelity across the pixel domain, the pixel-to-latent mapping process, and the latent domain.

In the pixel domain, different from previous methods \cite{wang2024exploiting,yang2025pixel,wu2024seesr,qu2024xpsr,chen2025faithdiff} that directly use bicubic upsampling to the target HR image resolution, we first use a lightweight pre-trained ×4 SR model \cite{zamfir2024see} for initial upsampling, followed by bicubic upsampling, providing a better initial input with better fidelity for the fidelity branch.

In the pixel-to-latent mapping process, there is a gap between the latent space of the blurry, coarsely upsampled LR images and the pre-trained high-quality image latent diffusion space. To align the latent features, we supervise the intermediate features of the mapping process, encouraging the LR features to approximate the high-quality HR features using a Latent Quality Loss (LQA). Specifically, similar to PASD \cite{yang2025pixel}, we apply convolution layers to convert the multi-scale intermediate image encoder features into corresponding-resolution RGB images and compute the $L_1$ distance between the converted RGB images and the downsampled corresponding-resolution ground-truth RGB images:
\begin{equation}
    L_{LQA} = \sum_{n=1,2,3}\|I_{HR}^{(n)} - toRGB(F_e^{(n)})\|
\end{equation}
where $F_e^{(n)}$ represents the $n$-th level image encoder intermediate features which are in 1/2, 1/4 and 1/8 scaled resolutions, $toRGB$ refers to the convolutional layers that convert the intermediate features into RGB images  (converting to a 3-channel RGB image without changing the resolution), and $I_{HR}^{(n)}$ represents the downsampled ground-truth HR that in the similar resolution of the $n$-th level intermediate features. 

In the latent space, to provide multi-grained fidelity control, we adopt a dual semantic-preserving mechanism that leverages both text prompts and fine-grained semantic feature conditions, in addition to the spatial control in the fidelity branch. Specifically, we utilize BLIP-2 \cite{li2023blip} to extract an image caption $c$ as the text prompt and use a fine-tuned semantic extractor RAM \cite{zhang2024recognize} to obtain fine-grained semantic features similar to SeeSR \cite{wu2024seesr}. As shown in Fig. \ref{fig:overview}, we place a Semantics-PReserving (SePR) attention after the text cross-attention layer to enforce the fine-grained semantic feature constraint. The SePR attention can be formulated as:
\begin{equation}
Attn(Q,K,V)=softmax(\frac{QK^T}{\sqrt{d_K}}) \cdot V
\end{equation}
where $d_K$ represents the dimension of the $K$ vector, and query $Q$, key $K$, value $V$ are obtained by:
\begin{equation}
    Q=W_Q^{(i)} \cdot F_{U}^{(i)}, K=W_K^{(i)} \cdot F_{sem}, V=W_V^{(i)} \cdot F_{sem}
\end{equation}
where $F_{U}$ denotes the $i$-th level denoising UNet features, $F_{sem}$ denotes the fine-grained semantic features, and $W_Q^{(i)}, W_K^{(i)}, W_V^{(i)}$ are learnable projection matrices.

\subsection{Loss Function and Training Strategy}
\subsubsection{Multi-resolution Training Strategy}
To enable ASSR for images of various sizes, we mix various-resolution HR images with sizes in the range of [32, 512] and their corresponding LR images with the SR scales ranging from 4 to 16. We first crop 512×512 patches and use bicubic sampling to obtain mixed-resolution HR-LR pairs with different SR scales.

\subsubsection{Loss Function}
During training, given a ground truth HR image, we first encode it using the pre-trained VAE encoder to obtain the latent representation $z_0$, and then progressively add noise to form the noisy latent $z_t$. Along with the LR image $I_{LR}$, SR scale $s$, diffusion step $t$, and the extracted dual-level semantics text prompt $c$ and semantic feature $p$, we train the diffusion denoising network $\epsilon_{\theta}$ of OmniScaleSR to predict the noise $\epsilon$ added to the noisy latent. The loss function is: 
\begin{equation}
    L_{diff} = \Bbb{E}_{z_0, t, c, p, I_{LR}, s, \epsilon \sim \cal{N}}[\|\epsilon - \epsilon_{\theta}(z_t, t, c, p, I_{LR}, s)\|_2^2]
\end{equation}

We jointly train the diffusion model and the convolution-based image encoder during training, and the overall loss is:
\begin{equation}
    L_{total} = L_{diff} + \alpha_{LQA}L_{LQA}
\end{equation}
where $\alpha_{LQA}$ is a hyperparameter. We empirically set $\alpha_{LQA}=1$ for the best practice. 

\begin{table*}[!htbp]
    \centering
    \renewcommand{\arraystretch}{0.65}
    \setlength{\tabcolsep}{6.6pt} % 调整列间距
    \caption{\sc ASSR-FINE-TUNING results of all FIXED-SR-SCALE-TRAINED baselines. Methods with ‘*’ denote the fine-tuned versions. IMPROVED metrics after fine-tuning are highlighted in \textcolor{blue}{\underline{blue}}, and the OVERALL BETTER version between the pre- and post-fine-tuned is highlighted in \colorbox{pink!70}{pink}.
    } 
    \begin{tabularx}{\textwidth}{c|c|cc|cc|cc|cc|cc}
    \toprule
     % &  & \multicolumn{2}{c|}{\textit{\textbf{DIV2K} (bicubic)}} & \multicolumn{10}{c}{\centering \textit{\textbf{RealSR} (Real-World Degradation)}} \\
     % \cmidrule{3-14}
    & Metrics & StableSR & \colorbox{pink!70}{StableSR$^*$} & PASD & \colorbox{pink!70}{PASD$^*$} & \colorbox{pink!70}{SeeSR} & SeeSR$^*$ & \colorbox{pink!70}{XPSR} & XPSR$^*$ & \colorbox{pink!70}{FaithDiff} & FaithDiff$^*$ \\
     \midrule
    \multirow{7}{*}{\parbox{20pt}{\centering ×5.3}}& PSNR$\uparrow$ & 22.43 & 21.45 & 21.63 & 19.81 & 21.22 & \textcolor{blue}{\underline{21.78}} & 19.53 & \textcolor{blue}{\underline{22.45}} & 21.08 & 21.42 \\
     & LPIPS$\downarrow$ & 0.2644 & 0.2959 & 0.3175 & 0.3509 & 0.3110 & \textcolor{blue}{\underline{0.2967}} & 0.3869 & \textcolor{blue}{\underline{0.3328}} & 0.2710 & 0.3285  \\
     & FID$\downarrow$ & 143.29 & 169.56 & 170.18 & 180.61 & 165.84 & \textcolor{blue}{\underline{158.71}} & 199.70 & \textcolor{blue}{\underline{166.42}} & 139.75 & 166.20  \\
     & DISTS$\downarrow$ & 0.2110 & 0.2439 & 0.2400 & 0.2706 & 0.2448 & \textcolor{blue}{\underline{0.2375}} & 0.2868 & \textcolor{blue}{\underline{0.2595}} & 0.2166 & 0.2485 \\
     & MANIQA$\uparrow$ & 0.4512 & \textcolor{blue}{\underline{0.5785}} & 0.4023 & \textcolor{blue}{\underline{0.6028}} & 0.5004 & \textcolor{blue}{\underline{0.6158}} & 0.5814 & \textcolor{blue}{\underline{0.5987}} & 0.4825 & \textcolor{blue}{\underline{0.5013}} \\
     & MUSIQ$\uparrow$ & 63.60 & \textcolor{blue}{\underline{68.57}} & 57.82 & \textcolor{blue}{\underline{67.88}} & 66.93 & \textcolor{blue}{\underline{69.29}} & 68.94 & 64.27 & 67.17 & 63.94 \\
     & CLIPIQA$\uparrow$ & 0.5886 & \textcolor{blue}{\underline{0.6279}} & 0.5577 & \textcolor{blue}{\underline{0.7464}} & 0.5824 & \textcolor{blue}{\underline{0.6785}} & 0.6420 & \textcolor{blue}{\underline{0.7128}} & 0.6169 & 0.5774 \\

    \cmidrule{1-12}
     \multirow{7}{*}{\parbox{20pt}{\centering ×10.7}}& PSNR$\uparrow$ & 21.23 & 20.62 & 19.52 & 18.06 & 18.88 & \textcolor{blue}{\underline{19.96}} & 18.54 & \textcolor{blue}{\underline{20.53}} & 20.03 & 19.29 \\
     & LPIPS$\downarrow$ & 0.4134 & \textcolor{blue}{\underline{0.4061}} & 0.4338 & 0.4490 & 0.4203 & \textcolor{blue}{\underline{0.3984}} & 0.4589 & \textcolor{blue}{\underline{0.4509}} & 0.3580 & 0.4029  \\
     & FID$\downarrow$ & 207.35 & 213.24 & 215.73 & 225.89 & 212.63 & \textcolor{blue}{\underline{189.23}} & 227.99 & \textcolor{blue}{\underline{213.08}} & 176.37 & 193.27 \\
     & DISTS$\downarrow$ & 0.2841 & 0.2906 & 0.2895 & 0.3050 & 0.2900 & \textcolor{blue}{\underline{0.2774}} & 0.3104 & \textcolor{blue}{\underline{0.3004}} & 0.2791 & 0.2823 \\
     & MANIQA$\uparrow$ & 0.3094 & \textcolor{blue}{\underline{0.4936}} & 0.3669 & \textcolor{blue}{\underline{0.6405}} & 0.4781 & \textcolor{blue}{\underline{0.6077}} & 0.5232 & \textcolor{blue}{\underline{0.6009}} & 0.4872 & \textcolor{blue}{\underline{0.4961}} \\
     & MUSIQ$\uparrow$ & 45.77 & \textcolor{blue}{\underline{61.74}} & 53.33 & \textcolor{blue}{\underline{65.57}} & 65.41 & \textcolor{blue}{\underline{67.85}} & 66.87 & 64.74 & 67.36 & 64.88 \\
     & CLIPIQA$\uparrow$ & 0.4270 & \textcolor{blue}{\underline{0.5289}} & 0.4899 & \textcolor{blue}{\underline{0.7314}} & 0.5272 & \textcolor{blue}{\underline{0.6266}} & 0.6114 & \textcolor{blue}{\underline{0.7162}} & 0.6157 & 0.5668 \\

     \cmidrule{1-12}
     \multirow{7}{*}{\parbox{20pt}{\centering ×16}}& PSNR$\uparrow$ & 21.64 & 20.59 & 20.44 & 18.94 & 19.66 & \textcolor{blue}{\underline{19.84}} & 19.53 & \textcolor{blue}{\underline{20.21}} & 19.55 & 18.75  \\
     & LPIPS$\downarrow$ & 0.5821 & \textcolor{blue}{\underline{0.4578}} & 0.5379 & \textcolor{blue}{\underline{0.5057}} & 0.4620 & \textcolor{blue}{\underline{0.4535}} & 0.5122 & 0.5128 & 0.4688 & 0.4777 \\
     & FID$\downarrow$ & 238.24 & \textcolor{blue}{\underline{208.66}} & 217.58 & 225.23 & 201.77 & \textcolor{blue}{\underline{198.83}} & 212.88 & 218.78 & 194.23 & 195.30  \\
     & DISTS$\downarrow$ & 0.3781 & \textcolor{blue}{\underline{0.2915}} & 0.3245 & \textcolor{blue}{\underline{0.3160}} & 0.2950 & 0.2961 & 0.3103 & 0.3139 & 0.2963 & 0.2972  \\
     & MANIQA$\uparrow$ & 0.2100 & \textcolor{blue}{\underline{0.4413}} & 0.3559 & \textcolor{blue}{\underline{0.5648}} & 0.5573 & 0.5165 & 0.6051 & 0.5492 & 0.4632 & 0.4346 \\
     & MUSIQ$\uparrow$ & 23.73 & \textcolor{blue}{\underline{61.54}} & 47.22 & \textcolor{blue}{\underline{64.08}} & 69.74 & 67.55 & 69.47 & 64.14 & 68.66 & 64.27 \\
     & CLIPIQA$\uparrow$ & 0.2770 & \textcolor{blue}{\underline{0.5468}} & 0.4267 & \textcolor{blue}{\underline{0.6661}} & 0.6699 & 0.5898 & 0.7390 & 0.7111 & 0.5874 & 0.5361 \\

     \cmidrule{1-12}
     \multirow{7}{*}{\parbox{20pt}{\centering ×20.8}}& PSNR$\uparrow$ & 20.76 & 19.96 & 20.55 & 18.73 & 19.02 & 18.86 & 19.06 & \textcolor{blue}{\underline{19.27}} & 18.81 & 18.57  \\
     & LPIPS$\downarrow$ & 0.7001 & \textcolor{blue}{\underline{0.5601}} & 0.6677 & \textcolor{blue}{\underline{0.5751}} & 0.5240 & 0.5251 & 0.5679 & 0.5680 & 0.5257 & \textcolor{blue}{\underline{0.5056}}   \\
     & FID$\downarrow$ & 231.81 & \textcolor{blue}{\underline{216.98}} & 227.55 & \textcolor{blue}{\underline{212.84}} & 193.78 & 204.37 & 209.46 & 216.66 & 193.08 & \textcolor{blue}{\underline{189.23}}  \\
     & DISTS$\downarrow$ & 0.4508 & \textcolor{blue}{\underline{0.3337}} & 0.4297 & \textcolor{blue}{\underline{0.3261}} & 0.3106 & 0.3112 & 0.3174 & 0.3299 & 0.2971 & \textcolor{blue}{\underline{0.2879}}  \\
     & MANIQA$\uparrow$ & 0.2423 & \textcolor{blue}{\underline{0.3517}} & 0.2759 & \textcolor{blue}{\underline{0.5073}} & 0.5377 & 0.4795 & 0.5803 & 0.4877 & 0.4703 & 0.3760 \\
     & MUSIQ$\uparrow$ & 18.18 & \textcolor{blue}{\underline{53.11}} & 24.46 & \textcolor{blue}{\underline{61.84}} & 67.70 & 67.67 & 68.41 & 61.49 & 70.10 & 61.57  \\
     & CLIPIQA$\uparrow$ & 0.2508 & \textcolor{blue}{\underline{0.4314}} & 0.2573 & \textcolor{blue}{\underline{0.5879}} & 0.6255 & 0.5235 & 0.7331 & 0.6641 & 0.6272 & 0.4940 \\

     \cmidrule{1-12}
     \multirow{7}{*}{\parbox{20pt}{\centering ×24}}& PSNR$\uparrow$ & 20.70 & 20.07 & 20.65 & 18.80 & 19.07 & 18.90 & 18.99 & \textcolor{blue}{\underline{19.40}} & 18.83 & 18.67  \\
     & LPIPS$\downarrow$ & 0.7159 & \textcolor{blue}{\underline{0.5950}} & 0.6975 & \textcolor{blue}{\underline{0.6088}} & 0.5438 & 0.5445 & 0.5860 & 0.5962 & 0.5746 & \textcolor{blue}{\underline{0.5321}} \\
     & FID$\downarrow$ & 231.76 & \textcolor{blue}{\underline{228.80}} & 228.55 & \textcolor{blue}{\underline{222.57}} & 209.17 & 211.23 & 211.82 & 226.30 & 200.16 & \textcolor{blue}{\underline{196.14}}   \\
     & DISTS$\downarrow$ & 0.4851 & \textcolor{blue}{\underline{0.3531}} & 0.4832 & \textcolor{blue}{\underline{0.3468}} & 0.3322 & 0.3367 & 0.3314 & 0.3626 & 0.3288 & \textcolor{blue}{\underline{0.2930}}   \\
     & MANIQA$\uparrow$ & 0.2742 & \textcolor{blue}{\underline{0.3053}} & 0.2860 & \textcolor{blue}{\underline{0.4978}} & 0.5373 & 0.4595 & 0.5671 & 0.4640 & 0.4620 & 0.3620  \\
     & MUSIQ$\uparrow$ & 16.52 & \textcolor{blue}{\underline{43.51}} & 18.20 & \textcolor{blue}{\underline{57.67}} & 66.57 & 65.66 & 67.62 & 54.37 & 65.99 & 58.67 \\
     & CLIPIQA$\uparrow$ & 0.2651 & \textcolor{blue}{\underline{0.3976}} & 0.2312 & \textcolor{blue}{\underline{0.5661}} & 0.6392 & 0.5033 & 0.7344 & 0.6367 & 0.5913 & 0.4696 \\
    
    \bottomrule
    \end{tabularx}
    \label{tab:tune}
    \vspace{-1em}
\end{table*}

\begin{table*}[!htbp]
    \centering
    \renewcommand{\arraystretch}{0.65}
    \setlength{\tabcolsep}{4.1pt} % 调整列间距
    \caption{\sc Quantitative comparison results on BICUBIC downsampling degradation datasets.
    We FINE-TUNED all FIXED-SR-SCALE-TRAINED SR baselines under ASSR settings for fairness, and report the BETTER VERSION to represent their best performance. The best and second-best results are highlighted in \textbf{\textcolor{red}{red}} and \textcolor{blue}{\underline{blue}}, respectively.
    } 
    \begin{tabularx}{\textwidth}{c|cc|ccccc|ccccccc}
    \toprule
    & & & \multicolumn{5}{c|}{\textbf{Non-Diffusion-based Methods}} & \multicolumn{6}{c}{\textbf{Diffusion-based Methods}} \\
    Datasets &  & Metrics & LIIF & CiaoSR & OPE-SR & LMF & GaussianSR & IDM & StableSR & PASD & SeeSR & XPSR & FaithDiff & Ours \\
     \midrule
     \multirow{40}{*}{\parbox{30pt}{\centering \textit{DIV2K}}}& \multirow{7}{*}{\parbox{12pt}{\centering ×5.3}} & PSNR$\uparrow$ & \textbf{\textcolor{red}{27.30}} & \textcolor{blue}{\underline{27.20}} & 26.84 & 26.84 & 26.62 & 19.12 & 21.69 & 20.72 & 23.33 & 20.25 & 22.56 & 22.10 \\
     &  & LPIPS$\downarrow$ & 0.2843 & 0.2998 & 0.3257 & 0.3195 & 0.3373 & 0.3378 & 0.2624 & 0.3098 & \textbf{\textcolor{red}{0.2193}} & 0.3515 & \textcolor{blue}{\underline{0.2199}} & 0.2325 \\
     &  & FID$\downarrow$ & 63.62 & 61.28 & 60.85 & 60.57 & 61.80 & 134.75 & 53.26 & 52.46 & \textcolor{blue}{\underline{33.72}} & 61.80 & \textbf{\textcolor{red}{32.54}} & 36.26 \\
     &  & DISTS$\downarrow$ & 0.2179 & 0.2230 & 0.2283 & 0.2291 & 0.2380 & 0.2822 & 0.2206 & 0.2290 & \underline{\textcolor{blue}{0.1763}} & 0.2509 & \textbf{\textcolor{red}{0.1725}} & 0.2004 \\
     &  & MANIQA$\uparrow$ & 0.3443 & 0.3589 & 0.3226 & 0.3259 & 0.3076 & 0.4225 & 0.5586 & \underline{\textcolor{blue}{0.5838}} & 0.4698 & 0.5468 & 0.4474 & \textbf{\textcolor{red}{0.5939}} \\
     &  & MUSIQ$\uparrow$ & 48.98 & 50.03 & 45.61 & 46.75 & 44.35 & 63.95 & \textcolor{blue}{\underline{67.63}} & 66.52 & 65.01 & 65.09 & 65.74 & \textbf{\textcolor{red}{67.67}} \\
     &  & CLIPIQA$\uparrow$ & 0.3953 & 0.4002 & 0.3929 & 0.3865 & 0.3904 & 0.6388 & 0.6944 & \textcolor{blue}{\underline{0.7035}} & 0.6310 & 0.6818 & 0.6574 & \textbf{\textcolor{red}{0.7476}} \\

    \cmidrule{2-15}
     & \multirow{7}{*}{\parbox{12pt}{\centering ×10.7}} & PSNR$\uparrow$ & 23.44 & \textbf{\textcolor{red}{23.55}} & \textcolor{blue}{\underline{23.48}} & \textcolor{blue}{\underline{23.48}} & 23.08 & 22.06 & 21.03 & 19.20 & 20.02 & 18.74 & 20.15 & 20.23 \\
     &  & LPIPS$\downarrow$ & 0.4938 & 0.3709 & 0.5239 & 0.5166 & 0.5609 & 0.4295 & 0.3944 & 0.4277 & 0.3676 & 0.4379 &\textbf{\textcolor{red}{0.3406}} & \textcolor{blue}{\underline{0.3605}} \\
     &  & FID$\downarrow$ & 110.35 & 105.42 & 102.30 & 103.98 & 102.20 & 130.58 & 70.06 & 78.28 & 54.51 & 81.88 & \textcolor{blue}{\underline{50.91}}& \textbf{\textcolor{red}{50.29}} \\
     &  & DISTS$\downarrow$ & 0.3234 & 0.3153 & 0.3264 & 0.3268 & 0.3430 & 0.2995 & 0.2713 & 0.2745 & 0.2428 & 0.2801 & \textbf{\textcolor{red}{0.2302}} & \textcolor{blue}{\underline{0.2423}} \\
     &  & MANIQA$\uparrow$ & 0.2457 & 0.2646 & 0.2337 & 0.2314 & 0.1859 & 0.2481 & 0.4208 & 0.4812 & 0.4812 & \textcolor{blue}{\underline{0.5262}} & 0.4520& \textbf{\textcolor{red}{0.6162}} \\
     &  & MUSIQ$\uparrow$ & 33.35 & 35.55 & 27.74 & 28.95 & 23.31 & 41.70 & 55.23 & 62.85 & 65.40 & \textcolor{blue}{\underline{66.06}} &66.04 & \textbf{\textcolor{red}{67.81}} \\
     &  & CLIPIQA$\uparrow$ & 0.2995 & 0.3060 & 0.3075 & 0.2889 & 0.2724 & 0.4392 & 0.5255 & \textcolor{blue}{\underline{0.7333}} & 0.6333 & 0.6792 & 0.6598& \textcolor{red}{\textbf{0.7561}} \\

     \cmidrule{2-15}
     & \multirow{7}{*}{\parbox{12pt}{\centering ×16}} & PSNR$\uparrow$ & 22.75 & \textbf{\textcolor{red}{22.97}} & \textcolor{blue}{\underline{22.79}} & \textcolor{blue}{\underline{22.79}} & 22.27 & 20.56 & 20.78 & 19.51 & 19.47 & 19.22 & 19.58 & 19.51 \\
     &  & LPIPS$\downarrow$ & 0.5859 & 0.5703 & 0.6168 & 0.6089 & 0.6628 & 0.5222 & 0.4501 & 0.5113 & 0.4414 & 0.4755 & \underline{\textcolor{blue}{0.4292}} & \textcolor{red}{\textbf{0.4288}} \\
     &  & FID$\downarrow$ & 111.17 & 107.13 & 103.31 & 104.71 & 104.21 & 171.60 & 64.51 & 78.43 & 54.79 & 57.20 & \underline{\textcolor{blue}{52.77}} & \textbf{\textcolor{red}{52.59}} \\
     &  & DISTS$\downarrow$ & 0.3604 & 0.3577 & 0.3652 & 0.3658 & 0.3904 & 0.3428 & 0.2724 & 0.2832 & 0.2519 & 0.2636 & \textcolor{blue}{\underline{0.2472}} & \textbf{\textcolor{red}{0.2439}} \\
     &  & MANIQA$\uparrow$ & 0.2265 & 0.2442 & 0.2291 & 0.2231 & 0.1986 & 0.2525 & 0.4598 & 0.4933 & 0.5268 & \textcolor{red}{\textbf{0.5972}} & 0.4349 & \underline{\textcolor{blue}{0.5622}} \\
     &  & MUSIQ$\uparrow$ & 30.46 & 33.50 & 24.95 & 26.03 & 19.32 & 45.35 & 54.02 & 61.97 & 69.05 & \textcolor{blue}{\underline{69.38}} & 68.59 & \textbf{\textcolor{red}{70.57}} \\
     &  & CLIPIQA$\uparrow$ & 0.2744 & 0.2821 & 0.2888 & 0.2710 & 0.2563 & 0.2722 & 0.5155 & 0.6714 & 0.7120 & \textbf{\textcolor{red}{0.7769}} & 0.6407 & \textcolor{blue}{\underline{0.7470}} \\

     \cmidrule{2-15}
     & \multirow{7}{*}{\parbox{12pt}{\centering ×20.8}} & PSNR$\uparrow$ & 21.13 & \textbf{\textcolor{red}{21.31}} & \textcolor{blue}{\underline{21.17}} & 21.16 & 20.68 & 19.29 & 17.52 & 18.60 & 18.66 & 18.58  & 16.65 & 19.58 \\
     &  & LPIPS$\downarrow$ & 0.6356 & 0.6130 & 0.6678 & 0.6588 & 0.7160 & 0.5705 & 0.5440 & 0.5420 & \textcolor{blue}{\underline{0.4592}} & 0.5147  & 0.4945 & \textbf{\textcolor{red}{0.4469}} \\
     &  & FID$\downarrow$ & 135.89 & 131.38 & 125.80 & 128.65 & 127.76 & 209.52 & 101.11 & 90.61 & 67.00 & 71.64 & \underline{\textcolor{blue}{67.68}} & \textcolor{red}{\textbf{66.56}} \\
     &  & DISTS$\downarrow$ & 0.3909 & 0.3863 & 0.3979 & 0.3980 & 0.4307 & 0.3818 & 0.3244 & 0.2918 & \textcolor{blue}{\underline{0.2506}} & 0.2714 & 0.2596& \textbf{\textcolor{red}{0.2417}} \\
     &  & MANIQA$\uparrow$ & 0.2118 & 0.2150 & 0.2221 & 0.2143 & 0.2077 & 0.2838 & 0.2728 & 0.4721 & 0.4234 & \textcolor{red}{\textbf{0.5915}} &0.3530 & \underline{\textcolor{blue}{0.5591}} \\
     &  & MUSIQ$\uparrow$ & 28.08 & 30.40 & 22.80 & 23.84 & 17.66 & 47.57 & 42.69 & 61.56 & 67.54 & \textcolor{blue}{\underline{69.17}} & 64.27 & \textbf{\textcolor{red}{70.73}} \\
     &  & CLIPIQA$\uparrow$ & 0.2661 & 0.2484 & 0.2882 & 0.2678 & 0.2509 & 0.2405 & 0.4027 & 0.6271 & 0.6109 & \textcolor{red}{\textbf{0.7795}} & 0.5273 & \underline{\textcolor{blue}{0.7551}} \\

     \cmidrule{2-15}
     & \multirow{7}{*}{\parbox{12pt}{\centering ×24}} & PSNR$\uparrow$ & 20.86 & \textbf{\textcolor{red}{21.02}} & \textcolor{blue}{\underline{20.89}} & 20.88 & 20.39 & 18.99 & 19.45 & 18.34 & 16.84 & 18.41 & 16.40 & 19.51 \\
     &  & LPIPS$\downarrow$ & 0.6678 & 0.6528 & 0.6989 & 0.6899 & 0.7454 & 0.6073 & 0.6066 & 0.5905 & \textcolor{blue}{\underline{0.5066}} & 0.5410 & 0.5519 & \textbf{\textcolor{red}{0.4736}} \\
     &  & FID$\downarrow$ & 136.01 & 132.81 & 125.97 & 128.74 & 128.51 & 221.99 & 113.42 & 95.31 & \textcolor{blue}{\underline{72.54}} & 78.53 & 76.22 & \textbf{\textcolor{red}{71.27}} \\
     &  & DISTS$\downarrow$ & 0.4072 & 0.4069 & 0.4160 & 0.4154 & 0.4523 & 0.4237 & 0.3578 & 0.3127 & 0.2788 & 0.2835 & \textcolor{blue}{\underline{0.2786}} & \textbf{\textcolor{red}{0.2519}} \\
     &  & MANIQA$\uparrow$ & 0.2199 & 0.2351 & 0.2352 & 0.2262 & 0.2288 & 0.3708 & 0.2576 & 0.4597 & 0.5058 & \textcolor{red}{\textbf{0.5781}} & 0.3349 & \underline{\textcolor{blue}{0.5292}} \\
     &  & MUSIQ$\uparrow$ & 26.46 & 29.56 & 21.42 & 0.2262 & 16.57 & 53.22 & 37.05 & 59.22 & 66.85 & \textcolor{blue}{\underline{68.03}} & 61.96 & \textbf{\textcolor{red}{69.15}} \\
     &  & CLIPIQA$\uparrow$ & 0.2794 & 0.2828 & 0.3127 & 0.2846 & 0.2679 & 0.3027 & 0.3823 & 0.6139 & 0.6723 & \textcolor{red}{\textbf{0.7802}} & 0.5112 & \underline{\textcolor{blue}{0.7440}} \\

     \midrule
     \midrule
     \multirow{40}{*}{\parbox{30pt}{\centering \textit{Urban}}} & \multirow{7}{*}{\parbox{12pt}{\centering ×5.3}} & PSNR$\uparrow$ & \textcolor{blue}{\underline{23.38}} & \textbf{\textcolor{red}{24.05}} & 23.33 & 23.24 & 22.88 & 18.97 & 20.52 & 20.68 & 21.23 & 18.95 & 20.78 & 20.62 \\
     &  & LPIPS$\downarrow$ & 0.2767 & 0.2501 & 0.2910 & 0.2998 & 0.3098 & 0.3047 & 0.2419 & 0.2212 & 0.2073 & 0.2637 & \textbf{\textcolor{red}{0.1752}} & \textcolor{blue}{\underline{0.1983}} \\
     &  & FID$\downarrow$ & \textcolor{blue}{\underline{15.59}} & \textbf{\textcolor{red}{13.13}} & 15.80 & 16.13 & 18.76 & 52.44 & 42.81 & 65.53 & 28.31 & 40.58 & 23.35 & 30.86 \\
     &  & DISTS$\downarrow$ & 0.2009 & 0.1909 & 0.2024 & 0.1985 & 0.2169 & 0.2591 & 0.1614 & 0.1444 & 0.1520 & 0.1825 & \textbf{\textcolor{red}{0.1334}} & \textcolor{blue}{\underline{0.1426}} \\
     &  & MANIQA$\uparrow$ & 0.4205 & 0.4426 & 0.3964 & 0.3904 & 0.3796 & 0.4097 & 0.5035 & 0.5973 & 0.6059 & \textbf{\textcolor{red}{0.7141}} & 0.5261 & \textcolor{blue}{\underline{0.6303}} \\
     &  & MUSIQ$\uparrow$ & 66.35 & 67.34 & 64.38 & 63.40 & 62.89 & 67.76 & 70.69 & 72.23 & 72.52 & \textcolor{blue}{\underline{73.31}} & 72.84&  \textbf{\textcolor{red}{73.44}} \\
     &  & CLIPIQA$\uparrow$ & 0.4410 & 0.4471 & 0.4233 & 0.4402 & 0.4202 & 0.6503 & 0.5719 & 0.7115 & 0.6858 & \textcolor{blue}{\underline{0.7615}} & 0.6533 & \textbf{\textcolor{red}{0.7239}} \\

     \cmidrule{2-15}
     & \multirow{7}{*}{\parbox{12pt}{\centering ×10.7}} & PSNR$\uparrow$ & \textcolor{blue}{\underline{20.58}} & \textbf{\textcolor{red}{21.02}} & 20.13 & 20.09 & 19.64 & 18.48 & 19.68 & 18.75 & 18.78 & 18.24 & 18.21 & 18.12 \\
     &  & LPIPS$\downarrow$ & 0.4576 & 0.4160 & 0.5154 & 0.5294 & 0.5774  & 0.5124 & 0.3182 & 0.3565 & 0.3050 & 0.3563 & \textbf{\textcolor{red}{0.2764}} & \textcolor{blue}{\underline{0.3096}} \\
     &  & FID$\downarrow$ & 87.46 & 78.92 & 87.96 & 89.44 & 100.35 & 131.83 & 70.04 & 72.26 & 61.03 & 78.05 & \textbf{\textcolor{red}{56.34}} & \textcolor{blue}{\underline{58.24}} \\
     &  & DISTS$\downarrow$ & 0.2955 & 0.2783 & 0.3101 & 0.3048 & 0.3440 & 0.3352 & 0.1929 & 0.2106 & 0.1981 & 0.2204 & \textcolor{blue}{\underline{0.1866}} & \textbf{\textcolor{red}{0.1816}} \\
     &  & MANIQA$\uparrow$ &  0.3055 & 0.3352 & 0.2859 & 0.2877 & 0.2250 & 0.2549 & 0.5309 & 0.5450 & 0.6040 & \textbf{\textcolor{red}{0.6683}} & 0.5117 & \textcolor{blue}{\underline{0.5634}} \\
     &  & MUSIQ$\uparrow$ & 51.67 & 55.75 & 46.71 & 44.83 & 38.72 & 50.51 & 70.54 & 69.50 & \textcolor{blue}{\underline{72.85}} & 73.14 & 72.77 & \textbf{\textcolor{red}{72.89}} \\
     &  & CLIPIQA$\uparrow$ & 0.3499 & 0.3550 & 0.3367 & 0.3633 & 0.2965 & 0.3997 & 0.6877 & 0.6768 & 0.6572 & \textbf{\textcolor{red}{0.7536}}  & 0.6405 & \textcolor{blue}{\underline{0.6980}} \\

     \cmidrule{2-15}
     & \multirow{7}{*}{\parbox{12pt}{\centering ×16}} & PSNR$\uparrow$ & \textcolor{blue}{\underline{18.90}} & \textbf{\textcolor{red}{19.10}} & 18.79 & 18.76 & 18.28 & 17.14 & 17.58 & 17.19 & 17.10 & 16.95 & 16.58 & 16.60 \\
     &  & LPIPS$\downarrow$ & 0.6031 & 0.5672 & 0.6422 & 0.6556 & 0.7190 & 0.6063 & 0.4788 & 0.4690 & 0.4036 & 0.4544  & \textcolor{blue}{\underline{0.4026}} & \textbf{\textcolor{red}{0.4021}} \\
     &  & FID$\downarrow$ & 144.98 & 134.41 & 146.72 & 144.80 & 164.64 & 208.48 & 114.91 & 110.41 & \textcolor{blue}{\underline{95.75}} & 114.06 & 100.64 & \textbf{\textcolor{red}{93.70}} \\
     &  & DISTS$\downarrow$ & 0.3638 & 0.3509 & 0.3718 & 0.3659 & 0.4139 & 0.3814 & 0.2749 & 0.2584 & 0.2443 & 0.2662 & \textcolor{blue}{\underline{0.2419}} & \textbf{\textcolor{red}{0.2177}} \\
     &  & MANIQA$\uparrow$ & 0.2553 & 0.2849 & 0.2491 & 0.2575 & 0.2007 & 0.3317 & 0.4007 & 0.5138 & 0.6041 & \textbf{\textcolor{red}{0.6449}} & 0.4794 & \textcolor{blue}{\underline{0.6131}} \\
     &  & MUSIQ$\uparrow$ & 42.46 & 46.32 & 37.41 & 35.57 & 27.37 & 52.89 & 61.29 & 67.36 & \textcolor{blue}{\underline{72.85}} & 71.74 & 72.08 & \textbf{\textcolor{red}{73.58}} \\
     &  & CLIPIQA$\uparrow$ & 0.3218 & 0.3274 & 0.3132 & 0.3475 & 0.2761 & 0.3156 & 0.5058 & 0.6678 & 0.6990 & \textbf{\textcolor{red}{0.7548}} & 0.6372 & \textcolor{blue}{\underline{0.7016}} \\

     \cmidrule{2-15}
     & \multirow{7}{*}{\parbox{12pt}{\centering ×20.8}} & PSNR$\uparrow$ & \textcolor{blue}{\underline{18.29}} & \textbf{\textcolor{red}{18.53}} & 18.02 & 18.01 & 17.52 & 16.58 & 16.86 & 16.53 & 16.55 & 16.37 & 16.10 & 17.12 \\
     &  & LPIPS$\downarrow$ & 0.6618 & 0.6166 & 0.7134 & 0.7262 & 0.7906 & 0.6498 & 0.6279 & 0.5736 & 0.5008 & 0.5412 & \textcolor{blue}{\underline{0.4750}} & \textbf{\textcolor{red}{0.4669}} \\
     &  & FID$\downarrow$ & 186.10 & 175.45 & 188.46 & 184.50 & 209.44 & 273.25 & 184.62 & 152.56 & \textcolor{red}{\textbf{124.32}} & 152.05 & 135.77 & \textcolor{blue}{\underline{132.64}} \\
     &  & DISTS$\downarrow$ & 0.3994 & 0.3842 & 0.4140 & 0.4080 & 0.4577 & 0.4150 & 0.3633 & 0.3195 & 0.2908 & 0.3111 & \textcolor{blue}{\underline{0.2813}} & \textbf{\textcolor{red}{0.2669}} \\
     &  & MANIQA$\uparrow$ & 0.2381 & 0.2536 & 0.2387 & 0.2509 & 0.2081 & 0.3932 & 0.5301 & 0.4774 & 0.4979 & \textbf{\textcolor{red}{0.6279}} & 0.4793 &  \textcolor{blue}{\underline{0.5888}} \\
     &  & MUSIQ$\uparrow$ & 35.69 & 39.30 & 30.94 & 29.42 & 21.95 & 55.17 & 56.01 & 60.98 & 69.00 & \textcolor{blue}{\underline{69.88}} & 70.33 & \textbf{\textcolor{red}{71.81}} \\
     &  & CLIPIQA$\uparrow$ & 0.3101 & 0.2928 & 0.3175 & 0.3604 & 0.2728 & 0.3450 & 0.3982 & 0.6039 & 0.6163 & \textcolor{red}{\textbf{0.7474}} & 0.6335 & \underline{\textcolor{blue}{0.7301}} \\

     \cmidrule{2-15}
     & \multirow{7}{*}{\parbox{12pt}{\centering ×24}} & PSNR$\uparrow$ & \textbf{\textcolor{red}{17.84}} & \textcolor{blue}{\underline{17.79}} & 17.64 & 17.62 & 17.15 & 16.29 & 16.51 & 15.97 & 16.14 & 15.97 & 16.03 & 16.82 \\
     &  & LPIPS$\downarrow$ & 0.6996 & 0.6786 & 0.7453 & 0.7590 & 0.8192 & 0.6726 & 0.6607 & 0.6148 & \textcolor{blue}{\underline{0.5597}} & 0.5899 & 0.5677 & \textbf{\textcolor{red}{0.5088}} \\
     &  & FID$\downarrow$ & 206.04 & 191.78 & 206.60 & 197.77 & 226.75 & 299.26 & 196.56 & 169.02 & \textcolor{blue}{\underline{155.12}} & 171.21 & 164.77 &  \textbf{\textcolor{red}{151.18}} \\
     &  & DISTS$\downarrow$ & 0.4233 & 0.4132 & 0.4352 & 0.4302 & 0.4802 & 0.4307 & 0.3827 & 0.3444 & 0.3281 & \textcolor{blue}{\underline{0.3299}} &  0.3269 & \textbf{\textcolor{red}{0.2869}} \\
     &  & MANIQA$\uparrow$ & 0.2530 & 0.2568 & 0.2429 & 0.2568 & 0.2206 & 0.4189 & 0.2999 & 0.4866 & 0.5718 & \textcolor{red}{\textbf{0.6267}} & 0.4690 & \underline{\textcolor{blue}{0.5789}} \\
     &  & MUSIQ$\uparrow$ & 32.71 & 36.93 & 28.07 & 26.68 & 20.24 & 58.18 & 44.83 & 59.89 & 67.67 & \textcolor{blue}{\underline{69.76}} & 65.05 & \textbf{\textcolor{red}{70.99}} \\
     &  & CLIPIQA$\uparrow$ & 0.3164 & 0.3152 & 0.3248 & 0.3772 & 0.2743 & 0.3626 & 0.4098 & 0.6125 & 0.6786 & \textcolor{red}{\textbf{0.7506}} & 0.5999 & \underline{\textcolor{blue}{0.7320}} \\

    \bottomrule
    \end{tabularx}
    \label{tab: simple}
    \vspace{-1em}
\end{table*}

\begin{table*}[!htbp]
    \centering
    \renewcommand{\arraystretch}{0.65}
    \setlength{\tabcolsep}{2.4pt} % 调整列间距
    \caption{\sc Quantitative comparison results on COMPLEX REAL-WORLD degradation datasets. We FINE-TUNED all FIXED-SR-SCALE-TRAINED SR baselines under ASSR settings for fairness, and report the BETTER VERSION to represent their best performance.
    The best and second-best results are highlighted in \textbf{\textcolor{red}{red}} and \textcolor{blue}{\underline{blue}}, respectively.
    } 
    \begin{tabularx}{\textwidth}{c|cc|cccccc|cc|cccccc}
    \toprule
    Datasets &  & Metrics & StableSR & PASD & SeeSR & XPSR & FaithDiff & Ours &  & Metrics & StableSR & PASD & SeeSR & XPSR & FaithDiff & Ours \\
     \midrule
     \multirow{26}{*}{\parbox{34pt}{\centering \textit{DIV2K\\(real)}}}& \multirow{7}{*}{\parbox{12pt}{\centering ×5.3}} & PSNR$\uparrow$ & 20.10 & 19.30 & 19.83 & 18.54 & \textbf{\textcolor{red}{20.33}} & \textcolor{blue}{\underline{20.22}}
     & \multirow{7}{*}{\parbox{12pt}{\centering ×16}} & PSNR$\uparrow$ & \textbf{\textcolor{red}{18.60}} & 17.73 & 17.78 & 17.60 & 17.71 & \textcolor{blue}{\underline{17.82}}  \\
     &  & LPIPS$\downarrow$ & 0.3065 & 0.3388 & 0.3211 & 0.3672 & \textbf{\textcolor{red}{0.2624}} & \textcolor{blue}{\underline{0.2903}} &  &
     LPIPS$\downarrow$ & 0.4723 & 0.5322 & 0.4600 & 0.4942 & \textcolor{blue}{\underline{0.4545}} & \textbf{\textcolor{red}{0.4497}} \\
     &  & FID$\downarrow$ & 55.70 & 56.55 & 51.46 & 65.96 & \underline{\textcolor{blue}{38.64}} & \textcolor{red}{\textbf{37.38}} &  &
     FID$\downarrow$ & 62.65 & 81.02 & 55.47 & 58.18 & \textcolor{blue}{\underline{54.97}} & \textbf{\textcolor{red}{51.52}} \\
     &  & DISTS$\downarrow$ & 0.2477 & 0.2380 & 0.2290 & 0.2572  & \textbf{\textcolor{red}{0.1991}} & \textcolor{blue}{\underline{0.2171}} &  &
     DISTS$\downarrow$ & 0.2923 & 0.2932 & \textcolor{blue}{\underline{0.2692}} & 0.2823 & 0.2710 & \textbf{\textcolor{red}{0.2593}} \\
     &  & MANIQA$\uparrow$ & 0.5357 & \textbf{\textcolor{red}{0.6243}} & 0.4693 & 0.5697 & 0.4487 & \textcolor{blue}{\underline{0.5980}} &  &
     MANIQA$\uparrow$ & 0.3693 & 0.4938 & 0.5251 & \textcolor{red}{\textbf{0.5962}} & 0.4376 & \underline{\textcolor{blue}{0.5618}} \\
     &  & MUSIQ$\uparrow$ & 66.00 & 65.17 & 64.74 & \textcolor{blue}{\underline{67.60}} & 65.70 & \textbf{\textcolor{red}{67.75}} &  &
     MUSIQ$\uparrow$ & 56.03 & 62.12 & 68.62 & \textcolor{blue}{\underline{69.24}} & 68.66 & \textbf{\textcolor{red}{70.52}} \\
     &  & CLIPIQA$\uparrow$ & 0.6700 & \textcolor{red}{\textbf{0.7624}} & 0.6389 & 0.7057 & 0.6002 & \underline{\textcolor{blue}{0.7521}} &  &
     CLIPIQA$\uparrow$ & 0.5296 & 0.6677 & 0.7074 & \textcolor{red}{\textbf{0.7764}} & 0.6433 & \underline{\textcolor{blue}{0.7595}} \\

    \cmidrule{2-17}
     & \multirow{7}{*}{\parbox{12pt}{\centering ×8}} & PSNR$\uparrow$ & \textbf{\textcolor{red}{20.41}} & 19.42 & \textcolor{blue}{\underline{19.93}} & 19.42 & 19.63 & 19.52
     & \multirow{7}{*}{\parbox{12pt}{\centering ×20.8}} & PSNR$\uparrow$ & \textcolor{blue}{\underline{17.64}} & 16.91 & 16.92 & 16.93 & 16.86 & \textbf{\textcolor{red}{17.77}} \\
     &  & LPIPS$\downarrow$ & 0.3619 & 0.4299 & 0.3587 & 0.3909  & \textbf{\textcolor{red}{0.3388}} & \textcolor{blue}{\underline{0.3575}} &  &
     LPIPS$\downarrow$ & 0.5667 & 0.5619 & 0.4930 & 0.5345 & \textcolor{blue}{\underline{0.4782}} & \textbf{\textcolor{red}{0.4659}} \\
     &  & FID$\downarrow$ & 42.15 & 55.62 & 39.33 & 42.23  & \textcolor{blue}{\underline{38.80}} & \textbf{\textcolor{red}{38.09}} &  &
     FID$\downarrow$ & 100.51 & 67.21 & 67.21 & 73.02 & \textbf{\textcolor{red}{62.83}} & \textcolor{blue}{\underline{65.76}} \\
     &  & DISTS$\downarrow$ & 0.2320 & 0.2438 & 0.2203 & 0.2376 & \textcolor{blue}{\underline{0.2201}} & \textbf{\textcolor{red}{0.2198}} &  &
     DISTS$\downarrow$ & 0.3446 & 0.3046 & 0.2824 & 0.2931  & \textcolor{blue}{\underline{0.2695}} & \textbf{\textcolor{red}{0.2621}} \\
     &  & MANIQA$\uparrow$ & 0.3930 & 0.5150 & 0.4930 & \textbf{\textcolor{red}{0.6013}}  & 0.4384 & \textcolor{blue}{\underline{0.5292}} &  &
     MANIQA$\uparrow$ & 0.2789 & 0.4686 & 0.5249 & \textcolor{blue}{\underline{0.5922}} & 0.4375 & \textbf{\textcolor{red}{0.5619}} \\
     &  & MUSIQ$\uparrow$ & 61.37 & 64.40 & 67.84 & \textcolor{blue}{\underline{69.55}} & 69.25 & \textbf{\textcolor{red}{69.66}} &  &
     MUSIQ$\uparrow$ & 44.37 & 61.70 & 68.96 & 69.60 & \textcolor{blue}{\underline{70.77}} & \textbf{\textcolor{red}{71.17}} \\
     &  & CLIPIQA$\uparrow$ & 0.5485 & 0.7028 & 0.6870 & \textbf{\textcolor{red}{0.7782}}  & 0.6517 & \textcolor{blue}{\underline{0.7335}} &  &
     CLIPIQA$\uparrow$ & 0.4098 & 0.6200 & 0.6901 & \textcolor{red}{\textbf{0.7811}} & 0.6614 & \underline{\textcolor{blue}{0.7592}} \\

     \cmidrule{2-17}
     & \multirow{7}{*}{\parbox{12pt}{\centering ×10.7}} & PSNR$\uparrow$ & \textbf{\textcolor{red}{19.06}} & 17.74 & 17.59 & 17.43 & 18.35 & \underline{\textcolor{blue}{18.63}}
     & \multirow{7}{*}{\parbox{12pt}{\centering ×24}} & PSNR$\uparrow$ & \textcolor{blue}{\underline{17.33}} & 16.57 & 16.68 & 16.68 & 16.65 & \textbf{\textcolor{red}{17.56}} \\
     &  & LPIPS$\downarrow$ & 0.4109 & 0.4426 & 0.4248 & 0.4543 & \textbf{\textcolor{red}{0.3672}} & \textcolor{blue}{\underline{0.3784}} &  &
     LPIPS$\downarrow$ & 0.6298 & 0.6119 & \textcolor{blue}{\underline{0.5344}} & 0.5625 & 0.5439 & \textbf{\textcolor{red}{0.4932}} \\
     &  & FID$\downarrow$ & 75.34 & 80.01 & 78.23 & 82.66 & \textcolor{blue}{\underline{52.50}} & \textbf{\textcolor{red}{48.93}} &  &
     FID$\downarrow$ & 114.71 & 94.92 & 74.83 & 84.52 & \textcolor{blue}{\underline{74.61}} & \textbf{\textcolor{red}{72.88}} \\
     &  & DISTS$\downarrow$ & 0.3023 & 0.2793 & 0.2814 & 0.2902 & \textbf{\textcolor{red}{0.2522}} & \textcolor{blue}{\underline{0.2541}} &  &
     DISTS$\downarrow$ & 0.3766 & 0.3265 & 0.3063 & \textcolor{blue}{\underline{0.3059}} & 0.3129 & \textbf{\textcolor{red}{0.2757}} \\
     &  & MANIQA$\uparrow$ & 0.4062 & \textcolor{blue}{\underline{0.5844}} & 0.4561 & 0.5203 & 0.4535 & \textbf{\textcolor{red}{0.6175}} &  &
     MANIQA$\uparrow$ & 0.2573 & 0.4611 & 0.5087 & \textbf{\textcolor{red}{0.5800}} & 0.4252 &\textcolor{blue}{\underline{0.5343}} \\
     &  & MUSIQ$\uparrow$ & 54.23 & 62.70 & 64.18 & 65.56 & \textcolor{blue}{\underline{65.98}} & \textbf{\textcolor{red}{68.11}} &  &
     MUSIQ$\uparrow$ & 38.31 & 59.42 & 66.99 & \textcolor{blue}{\underline{68.44}} & 65.12 & \textbf{\textcolor{red}{69.80}} \\
     &  & CLIPIQA$\uparrow$ & 0.5059 & \underline{\textcolor{blue}{0.7323}} & 0.5927 & 0.6699 & 0.6598 & \textcolor{red}{\textbf{0.7574}} &  &
     CLIPIQA$\uparrow$ & 0.3837 & 0.6064 & 0.6731 & \textcolor{red}{\textbf{0.7852}} & 0.6267 & \underline{\textcolor{blue}{0.7481}} \\

     \midrule
     \midrule
     \multirow{26}{*}{\parbox{34pt}{\centering \textit{RealSR}}} & \multirow{7}{*}{\parbox{12pt}{\centering ×5.3}} & PSNR$\uparrow$ & 21.45 & 19.81 & 21.22 & 19.53 & \textcolor{blue}{\underline{21.08}} &  \textbf{\textcolor{red}{21.51}} & 
     \multirow{7}{*}{\parbox{12pt}{\centering ×16}} & PSNR$\uparrow$ & \textbf{\textcolor{red}{20.59}} & 18.94 & 19.66 & 19.53 & 19.55 & \textcolor{blue}{\underline{19.72}} \\
     &  & LPIPS$\downarrow$ & \textcolor{blue}{\underline{0.2959}} & 0.3509 & 0.3110 & 0.3869 & \textbf{\textcolor{red}{0.2710}} & 0.3020 &  &
     LPIPS$\downarrow$ & \textcolor{blue}{\underline{0.4578}} & 0.5057 & 0.4620 & 0.5122 & 0.4688 & \textbf{\textcolor{red}{0.4569}} \\
     &  & FID$\downarrow$ & 169.56 & 180.61 & 165.84 & 199.70 & \textbf{\textcolor{red}{139.75}} & \textcolor{blue}{\underline{147.86}} &  &
     FID$\downarrow$ & 208.66 & 225.23 & 201.77 & 212.88 & \textcolor{blue}{\underline{194.23}} & \textbf{\textcolor{red}{181.41}} \\
     &  & DISTS$\downarrow$ & 0.2439 & 0.2706 & 0.2448 & 0.2868 & \textbf{\textcolor{red}{0.2166}} & \textcolor{blue}{\underline{0.2416}} &  &
     DISTS$\downarrow$ & \textcolor{blue}{\underline{0.2915}} & 0.3160 & 0.2950 & 0.3103 & 0.2963 & \textbf{\textcolor{red}{0.2862}} \\
     &  & MANIQA$\uparrow$ & 0.5785 & \textcolor{blue}{\underline{0.6028}} & 0.5004 & 0.5814 & 0.4825 & \textbf{\textcolor{red}{0.6247}} &  &
     MANIQA$\uparrow$ & 0.4413 & 0.5648 & 0.5573 & \textbf{\textcolor{red}{0.6051}} & 0.4632 & \textcolor{blue}{\underline{0.5915}} \\
     &  & MUSIQ$\uparrow$ & 68.57 & 67.88 & 66.93 & \textcolor{blue}{\underline{68.94}} & 67.17 & \textbf{\textcolor{red}{69.44}} &  &
     MUSIQ$\uparrow$ & 61.54 & 64.08 & \textcolor{blue}{\underline{69.74}} & 69.47 & 68.66 & \textbf{\textcolor{red}{70.97}} \\
     &  & CLIPIQA$\uparrow$ & 0.6279 & \textbf{\textcolor{red}{0.7464}} & 0.5824 & 0.6420 & 0.6169 & \textcolor{blue}{\underline{0.6985}} &  &
     CLIPIQA$\uparrow$ & 0.5468 & 0.6661 & 0.6699 & \textbf{\textcolor{red}{0.7390}} & 0.5874 & \textcolor{blue}{\underline{0.7106}} \\

     \cmidrule{2-17}
     & \multirow{7}{*}{\parbox{12pt}{\centering ×8}} & PSNR$\uparrow$ & \textbf{\textcolor{red}{22.56}} & 21.03 & 21.96 & \textcolor{blue}{\underline{21.55}} & 22.07 & 21.45
     & \multirow{7}{*}{\parbox{12pt}{\centering ×20.8}} & PSNR$\uparrow$ & \textcolor{blue}{\underline{19.96}} & 18.73 & 19.02 & 19.06 & 18.81 & \textbf{\textcolor{red}{20.05}} \\
     &  & LPIPS$\downarrow$ & \textcolor{blue}{\underline{0.3359}} & 0.3754 & 0.3464 & 0.4028 & \textbf{\textcolor{red}{0.3356}} & 0.3602 &  &
     LPIPS$\downarrow$ & 0.5601 & 0.5751 & \textcolor{blue}{\underline{0.5240}} & 0.5679 & 0.5257 & \textbf{\textcolor{red}{0.4957}} \\
     &  & FID$\downarrow$ & 163.80 & 169.86 & 149.57 & 163.43 & \textcolor{blue}{\underline{144.10}} & \textbf{\textcolor{red}{143.22}} &  &
     FID$\downarrow$ & 216.98 & 212.84 & 193.78 & 209.46 & \textcolor{blue}{\underline{193.08}} & \textbf{\textcolor{red}{192.19}} \\
     &  & DISTS$\downarrow$ &0.2485 & 0.2689 &  \textbf{\textcolor{red}{0.2440}} & 0.2708 & 0.2574 & \textcolor{blue}{\underline{0.2478}} &  &
     DISTS$\downarrow$ & 0.3337 & 0.3261 & 0.3106 & 0.3174 & \textcolor{blue}{\underline{0.2971}} & \textbf{\textcolor{red}{0.2936}} \\
     &  & MANIQA$\uparrow$ & 0.4456 & 0.5420 & 0.5375 & \textbf{\textcolor{red}{0.6179}} & 0.4772 & \textcolor{blue}{\underline{0.5575}} &  &
     MANIQA$\uparrow$ & 0.3517 & 0.5073 & 0.5377 & \textbf{\textcolor{red}{0.5803}} & 0.4703 & \textcolor{blue}{\underline{0.5549}} \\
     &  & MUSIQ$\uparrow$ & 63.63 & 66.77 & 69.18 & \textcolor{blue}{\underline{70.59}} & 70.14 & \textbf{\textcolor{red}{70.61}} &  &
     MUSIQ$\uparrow$ & 53.11 & 61.84 & 67.70 & \textcolor{blue}{\underline{68.41}} & 70.10 & \textbf{\textcolor{red}{68.85}} \\
     &  & CLIPIQA$\uparrow$ & 0.5379 & 0.6768 & 0.6527 & \textbf{\textcolor{red}{0.7405}} & 0.6085 & \textcolor{blue}{\underline{0.6881}} &  &
     CLIPIQA$\uparrow$ & 0.4314 & 0.5879 & 0.6255 & \textcolor{blue}{\underline{0.7331}} & 0.6272 & \textbf{\textcolor{red}{0.6904}} \\

     \cmidrule{2-17}
     & \multirow{7}{*}{\parbox{12pt}{\centering ×10.7}} & PSNR$\uparrow$ & \textbf{\textcolor{red}{20.62}} & 18.06 & 18.88 & 18.54 & \textcolor{blue}{\underline{20.03}} & 19.97
     & \multirow{7}{*}{\parbox{12pt}{\centering ×24}} & PSNR$\uparrow$ & \textcolor{blue}{\underline{20.07}} & 18.80 & 19.07 & 18.99 & 18.83 & \textbf{\textcolor{red}{20.29}} \\
     &  & LPIPS$\downarrow$ & 0.4061 & 0.4490 & 0.4203 & 0.4589 & \textcolor{blue}{\underline{0.3580}} & \textbf{\textcolor{red}{0.3954}} &  &
     LPIPS$\downarrow$ & 0.5950 & 0.6088 & \textcolor{blue}{\underline{0.5438}} & 0.5860 & 0.5746 & \textbf{\textcolor{red}{0.5104}} \\
     &  & FID$\downarrow$ & 213.24 & 225.89 & 212.63 & 227.99 & \textcolor{blue}{\underline{176.37}} &  \textbf{\textcolor{red}{171.94}} &  &
     FID$\downarrow$ & 228.80 & 222.57 & 209.17 & 211.82 & \textbf{\textcolor{red}{200.16}} & \textcolor{blue}{\underline{206.24}} \\
     &  & DISTS$\downarrow$ & 0.2906 & 0.3050 & 0.2900 & 0.3104 & \textcolor{blue}{\underline{0.2791}} & \textbf{\textcolor{red}{0.2741}} &  &
     DISTS$\downarrow$ & 0.3531 & 0.3468 & 0.3322 & 0.3314 & \textcolor{blue}{\underline{0.3288}} & \textbf{\textcolor{red}{0.3041}} \\
     &  & MANIQA$\uparrow$ & 0.4936 & \textcolor{blue}{\underline{0.6405}} & 0.4781 & 0.5232 & 0.4872 & \textbf{\textcolor{red}{0.6414}} &  &
     MANIQA$\uparrow$ & 0.3053 & 0.4978 & 0.5373 & \textbf{\textcolor{red}{0.5671}} & 0.4620 & \textcolor{blue}{\underline{0.5385}} \\
     &  & MUSIQ$\uparrow$ & 61.74 & 65.57 & 65.41 & 66.87 & \textcolor{blue}{\underline{67.36}} & \textbf{\textcolor{red}{69.49}} &  &
     MUSIQ$\uparrow$ & 43.51 & 57.67 & 66.57 & \textcolor{blue}{\underline{67.62}} & 65.99 & \textbf{\textcolor{red}{67.72}} \\
     &  & CLIPIQA$\uparrow$ & 0.5289 & \textbf{\textcolor{red}{0.7314}} & 0.5272 & 0.6114 & 0.6157 & \textcolor{blue}{\underline{0.7051}} &  &
     CLIPIQA$\uparrow$ & 0.3976 & 0.5661 & 0.6392 & \textcolor{blue}{\underline{0.7344}} & 0.5913 & \textbf{\textcolor{red}{0.6880}} \\

     \midrule
     \midrule
     \multirow{26}{*}{\parbox{34pt}{\centering \textit{DRealSR}}} & \multirow{7}{*}{\parbox{12pt}{\centering ×5.3}} & PSNR$\uparrow$ & 23.72 & 21.58 & 23.20 & 22.11 & \textcolor{blue}{\underline{23.71}} & \textbf{\textcolor{red}{23.73}}
     & \multirow{7}{*}{\parbox{12pt}{\centering ×16}} & PSNR$\uparrow$ & \textbf{\textcolor{red}{24.04}} & 22.23 & \textcolor{blue}{\underline{22.53}} & 22.41 & 22.45 & 22.40 \\
     &  & LPIPS$\downarrow$ & 0.3170 & 0.3536 & 0.3266 & 0.3655 & \textbf{\textcolor{red}{0.2808}} & \textcolor{blue}{\underline{0.3127}} &  &
     LPIPS$\downarrow$ & \textcolor{blue}{\underline{0.4349}} & 0.4840 & \textbf{\textcolor{red}{0.4333}} & 0.4801 & 0.4701 & 0.4593 \\
     &  & FID$\downarrow$ & 176.90 & 191.95 & 182.98 & 211.11 & \textcolor{blue}{\underline{166.21}} & \textbf{\textcolor{red}{161.83}} &  &
     FID$\downarrow$ & 201.06 & 229.12 & \textcolor{blue}{\underline{194.28}} & 205.27 & 208.69 & \textbf{\textcolor{red}{194.00}} \\
     &  & DISTS$\downarrow$ & 0.2488 & 0.2681 & 0.2558 & 0.2780 & \textbf{\textcolor{red}{0.2229}} & \textcolor{blue}{\underline{0.2472}} &  &
     DISTS$\downarrow$ & 0.2952 & 0.3124 & \textbf{\textcolor{red}{0.2855}} & 0.3037 & 0.3028 & \textcolor{blue}{\underline{0.2904}} \\
     &  & MANIQA$\uparrow$ & 0.5329 & \textbf{\textcolor{red}{0.6024}} & 0.4854 & 0.5498 & 0.4720 & \textcolor{blue}{\underline{0.5924}} &  &
     MANIQA$\uparrow$ & 0.3827 & 0.5038 & 0.5083 & \textcolor{blue}{\underline{0.5522}} & 0.4675 & \textbf{\textcolor{red}{0.5565}} \\
     &  & MUSIQ$\uparrow$ & 63.60 & 63.75 & 63.61 & \textcolor{blue}{\underline{65.82}} & 63.46 & \textbf{\textcolor{red}{66.11}} &  &
     MUSIQ$\uparrow$ & 53.74 & 59.04 & 65.18 & 65.52 & \textcolor{blue}{\underline{66.11}} & \textbf{\textcolor{red}{67.43}} \\
     &  & CLIPIQA$\uparrow$ & 0.6369 & \textbf{\textcolor{red}{0.7266}} & 0.6059 & 0.6755 & 0.6344 & \textcolor{blue}{\underline{0.7147}} &  &
     CLIPIQA$\uparrow$ & 0.5200 & 0.6411 & 0.6669 & \textbf{\textcolor{red}{0.7165}} & 0.6274 & \textcolor{blue}{\underline{0.7111}} \\

     \cmidrule{2-17}
     & \multirow{7}{*}{\parbox{12pt}{\centering ×8}} & PSNR$\uparrow$ & \textbf{\textcolor{red}{24.83}} & 23.28 &  \textcolor{blue}{\underline{24.02}} &23.48 & 23.58 & 23.38
     & \multirow{7}{*}{\parbox{12pt}{\centering ×20.8}} & PSNR$\uparrow$ & \textbf{\textcolor{red}{22.84}} & 21.61 & 21.62 & 21.61 & 21.24 & \textcolor{blue}{\underline{22.62}} \\
     &  & LPIPS$\downarrow$ & \textbf{\textcolor{red}{0.3384}} & 0.3894 & \textcolor{blue}{\underline{0.3433}} & 0.4012 & 0.3706 & 0.4006 &  &
     LPIPS$\downarrow$ & 0.5169 & 0.5513 & \textcolor{blue}{\underline{0.4783}} & 0.5220 & 0.5321 & \textbf{\textcolor{red}{0.4646}} \\
     &  & FID$\downarrow$ & 165.41 & 181.29 & 169.37 & 173.04 & \textcolor{blue}{\underline{175.37}} & \textbf{\textcolor{red}{168.48}} &  &
     FID$\downarrow$ & 208.94 & 209.29 & 190.69 & 198.61 & \textcolor{blue}{\underline{186.30}} & \textbf{\textcolor{red}{182.30}} \\
     &  & DISTS$\downarrow$ & 0.2711 & 0.2690 & 0.2733 & 0.2682 & \textcolor{blue}{\underline{0.2671}} & \textbf{\textcolor{red}{0.2657}} &  &
     DISTS$\downarrow$ & 0.3184 & 0.3389 & \textcolor{blue}{\underline{0.3026}} & 0.3141 & 0.3067 & \textbf{\textcolor{red}{0.2856}} \\
     &  & MANIQA$\uparrow$ & 0.3936 & 0.4837 & 0.5042 & \textbf{\textcolor{red}{0.5597}} & 0.4778 & \textcolor{blue}{\underline{0.5383}} &  &
     MANIQA$\uparrow$ & 0.3179 & 0.4608 & 0.5044 & \textcolor{blue}{\underline{0.5315}} & 0.4613 & \textbf{\textcolor{red}{0.5422}} \\
     &  & MUSIQ$\uparrow$ & 56.48 & 59.18 & 65.34 & 66.72 & \textcolor{blue}{\underline{67.29}} & \textbf{\textcolor{red}{67.75}} &  &
     MUSIQ$\uparrow$ & 47.22 & 55.15 & 65.15 & 64.89 & \textcolor{blue}{\underline{66.79}} & \textbf{\textcolor{red}{67.51}} \\
     &  & CLIPIQA$\uparrow$ & 0.5405 & 0.6649 & 0.6730 & \textbf{\textcolor{red}{0.7396}} & 0.6439 & \textcolor{blue}{\underline{0.7018}} &  &
     CLIPIQA$\uparrow$ & 0.4459 & 0.5761 & 0.6525 & \textcolor{blue}{\underline{0.7108}} & 0.6206 & \textbf{\textcolor{red}{0.7279}} \\

     \cmidrule{2-17}
     & \multirow{7}{*}{\parbox{12pt}{\centering ×10.7}} & PSNR$\uparrow$ & \textbf{\textcolor{red}{23.74}} & 20.77 & 21.50 & 21.50 & \textcolor{blue}{\underline{22.79}} & 22.66
     & \multirow{7}{*}{\parbox{12pt}{\centering ×24}} & PSNR$\uparrow$ & \textcolor{blue}{\underline{22.89}} & 21.85 & 21.84 & 21.57 & 21.37 & \textbf{\textcolor{red}{22.90}} \\
     &  & LPIPS$\downarrow$ & 0.3816 & 0.4332 & 0.4054 & 0.4269 & \textbf{\textcolor{red}{0.3514}} & \textcolor{blue}{\underline{0.3804}} &  &
     LPIPS$\downarrow$ & 0.5486 & 0.5702 & \textcolor{blue}{\underline{0.5160}} & 0.5404 & 0.5919 & \textbf{\textcolor{red}{0.4747}} \\
     &  & FID$\downarrow$ & 214.44 & 236.57 & 218.57 & 233.37 & \textcolor{blue}{\underline{191.55}} & \textbf{\textcolor{red}{189.86}} &  &
     FID$\downarrow$ & 224.57 & 212.70 & \textcolor{blue}{\underline{199.19}} & 206.95 & 199.68 & \textbf{\textcolor{red}{187.10}} \\
     &  & DISTS$\downarrow$ & 0.2825 & 0.3022 & 0.2929 & 0.3030 & \textbf{\textcolor{red}{0.2608}} & \textcolor{blue}{\underline{0.2739}} &  &
     DISTS$\downarrow$ & 0.3388 & 0.3624 & 0.3249 & \textcolor{blue}{\underline{0.3242}} & 0.3481 & \textbf{\textcolor{red}{0.2931}} \\
     &  & MANIQA$\uparrow$ & 0.4121 & \textcolor{blue}{\underline{0.5909}} & 0.4680 & 0.4987 & 0.4689 & \textbf{\textcolor{red}{0.6010}} &  &
     MANIQA$\uparrow$ & 0.2968 & 0.4410 & 0.4940 & \textbf{\textcolor{red}{0.5336}} & 0.4417 & \textcolor{blue}{\underline{0.5111}} \\
     &  & MUSIQ$\uparrow$ & 53.51 & 62.51 & 61.37 & 62.55 & \textcolor{blue}{\underline{63.97}} & \textbf{\textcolor{red}{66.17}} &  &
     MUSIQ$\uparrow$ & 42.35 & 51.83 & \textcolor{blue}{\underline{63.66}} & 63.44 & 59.83 & \textbf{\textcolor{red}{65.40}} \\
     &  & CLIPIQA$\uparrow$ & 0.4987 & \textbf{\textcolor{red}{0.7208}} & 0.5545 & 0.6236 & 0.6297 & \textcolor{blue}{\underline{0.7081}} &  &
     CLIPIQA$\uparrow$ & 0.4144 & 0.5615 & 0.6440 & \textbf{\textcolor{red}{0.7320}} & 0.5905 & \textcolor{blue}{\underline{0.7268}} \\

    \bottomrule
    \end{tabularx}
    \label{tab: real-world}
    \vspace{-1em}
\end{table*}

\section{Experiments}
\subsection{Experimental Setup}

\subsubsection{Training Datasets and Implementation Details}
We train on LSDIR \cite{li2023lsdir}, DIV2K \cite{agustsson2017ntire}, and the first 10k images of FFHQ dataset \cite{karras2019style}. We use the degradation pipeline of Real-ESRGAN \cite{wang2021realesrgan} to synthesize arbitrary-scale LR-HR training pairs with both bicubic and real-world downsampling degradations.
OminiScaleSR is built upon the pre-trained SD-2-base \cite{rombach2022high} model. All experiments are conducted on a single NVIDIA A100 GPU using PyTorch, with a batch size of 12. We apply the AdamW optimizer \cite{wang2004image} with a learning rate of 5e-5 and train the unfrozen parameters for 13K iterations, taking approximately 2 days. For inference, we adopt DDPM \cite{ho2020denoising} sampling with 50 timesteps.

\subsubsection{Evaluation Setting} \label{sec:setting}

We evaluate our method under two degradation settings: 
(1) \textbf{bicubic downsampling}, a widely used ideal degradation model; and
(2) \textbf{real-world degradation}, a more realistic degradation model involving complex and unknown distortions. We follow previous work \textcolor{blue}{and set} up two scenarios: (a) synthetic dataset using Real-ESRGAN \cite{wang2021realesrgan} data pipeline; (b) real-captured datasets. These datasets contain intricate distortions and are then subjected to scale-specific downsampling to implement various SR scale settings.

Correspondingly, we conduct evaluation on multiple evaluation datasets. For bicubic downsampling, we evaluate on DIV2K \cite{agustsson2017ntire} and Urban100 \cite{huang2015single}. For real-world degradation, we evaluate on synthetic DIV2K, and two real-captured datasets: DRealSR \cite{wei2020cdc} and RealSR \cite{cai2019toward}. We evaluate various SR scales, including ×5.3 (64→340), ×8 (64→512), ×10.7 (32→342), ×16 (32→512), ×20.8 (32→666) and ×24 (32→768). For example, “64→340” indicates super-resolving a 64×64 LR patch to a 340×340 HR patch.
Following prior works, we randomly crop 2,000 patches for DIV2K evaluation and apply center cropping for DRealSR and RealSR. The crop size is set to 512×512 for SR scales no more than ×16, and 768×768 for larger SR scales. For Urban100, we retain the original image resolution as HR and generate corresponding LR images at various SR scales to enable comprehensive evaluation across diverse resolutions.

\subsubsection{Metrics} To provide a comprehensive quantitative evaluation, we adopt a range of widely used reference and non-reference metrics, including PSNR, LPIPS \cite{zhang2018unreasonable} FID \cite{heusel2017gans}, DISTS \cite{ding2020image}, MANIQA \cite{yang2022maniqa}, MUSIQ \cite{ke2021musiq}, and CLIPIQA \cite{wang2023exploring}. For reference metrics, PSNR measures pixel-level similarity, while LPIPS and DISTS assess perceptual similarity. FID calculates the distribution distance between the ground truth and the generated images. For non-reference metrics, MANIQA assesses image quality using a multi-dimensional attention mechanism, MUSIQ captures multi-scale image representations to measure image quality at different granularities, while CLIPIQA leverages the pre-trained vision-language model CLIP to estimate perceptual quality.

\begin{figure*}[h!] % [h] 表示图片将尽可能放置在当前位置（here）
    \centering % 图片居中
    \includegraphics[width=\textwidth]{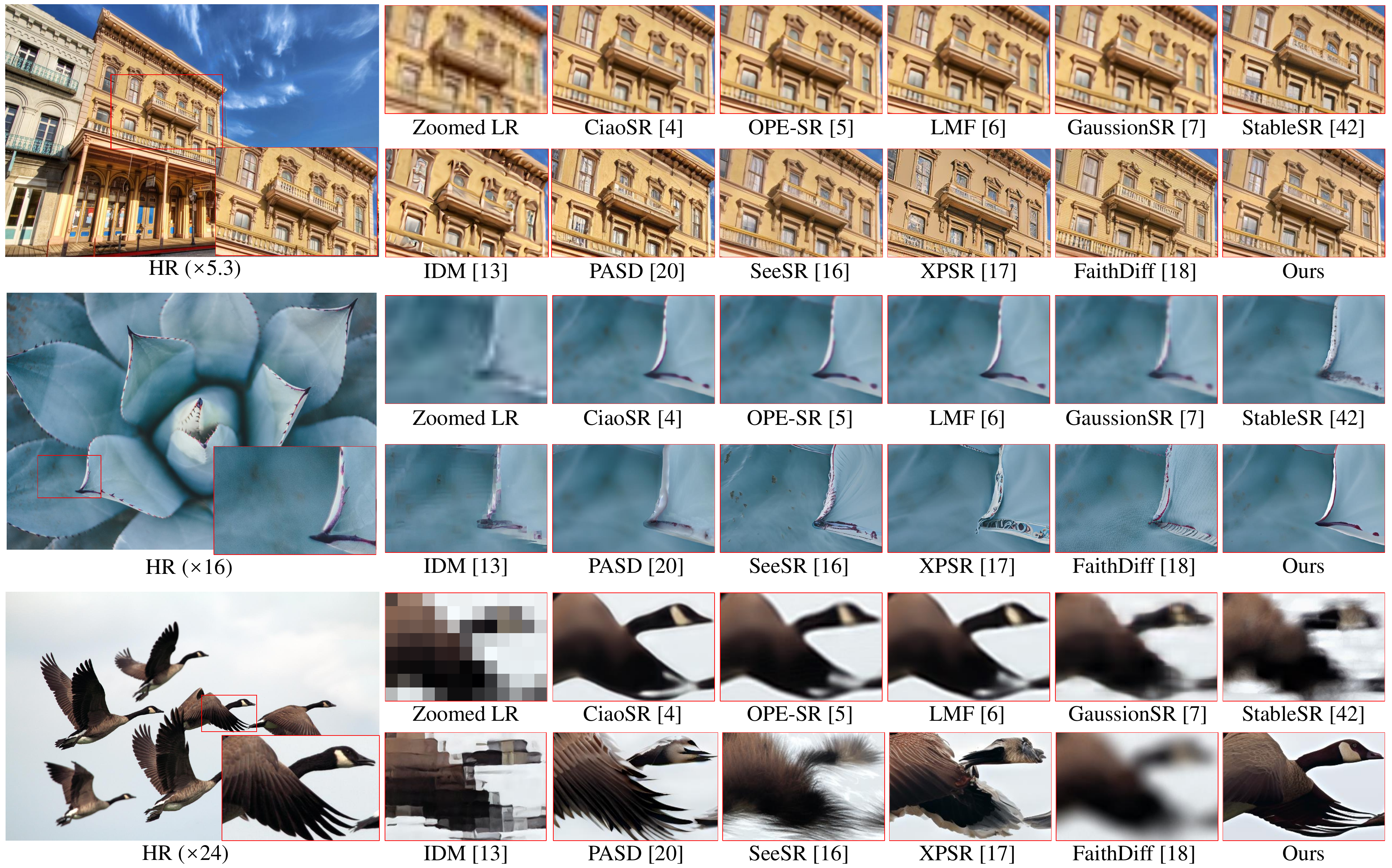} % 调整图片宽度为页面宽度的80%
    \caption{Qualitative comparisons under bicubic downsampling. SR scales from top to bottom: ×5.3, ×16, ×24.} % 添加图片标题
    \label{fig:clean} % 添加图片标签，便于引用
    % \vspace{-1em}
\end{figure*}

\subsubsection{Compared Methods} \label{sec:compared}
We compare our method against three categories of representative state-of-the-art methods, including (i) \textbf{Non-diffusion-based ASSR methods:} LIIF \cite{chen2021learning}, CiaoSR \cite{cao2023ciaosr}, OPE-SR \cite{song2023ope}, LMF \cite{he2024latent} and GaussianSR \cite{hu2025gaussiansr}, which incorporate INR into CNN, Transformer, or Gaussian Splatting architectures. (ii) \textbf{Regression-optimized Diffusion-based ASSR method:} IDM \cite{gao2023implicit}, which integrates INR into a pixel-domain diffusion framework. (iii) \textbf{Diffusion-based Real-ISR methods:} StableSR \cite{wang2024exploiting}, PASD \cite{yang2025pixel}, SeeSR \cite{wu2024seesr}, XPSR \cite{qu2024xpsr}, FaithDiff \cite{chen2025faithdiff}, which leverage diffusion prior but lack explicit SR scale controls.

Note that the non-diffusion-based ASSR methods \cite{chen2021learning,cao2023ciaosr,song2023ope,he2024latent,hu2025gaussiansr} and the regression-optimized diffusion-based method IDM \cite{gao2023implicit} are only evaluated under bicubic degradation, as they are organizationally designed for ideal synthetic degradation and cannot generalize to complex real-world degradations.

To ensure fair comparison, we have fine-tuned all fixed-SR-scale-trained baselines under ASSR training settings, including StableSR \cite{wang2024exploiting}, PASD \cite{yang2025pixel}, SeeSR \cite{wu2024seesr}, XPSR \cite{qu2024xpsr} and FaithDiff \cite{chen2025faithdiff}. We report their results before and after fine-tuning on RealSR dataset \cite{cai2019toward} in Tab. \ref{tab:tune}.
The experimental results show that \textbf{only the ASSR fine-tuning without explicit SR scale controls fails to bring new gains to the most SOTA Real-ISR method}, which further illustrates the necessity of explicit scale controls. We adopt \textbf{the better-performed version before and after fine-tuning to represent its best performance for fair comparison}. Specifically, we adopt the fine-tuned versions of StableSR and PASD, while using the official weights for SeeSR, XPSR, and FaithDiff, as their fine-tuned models do not yield better overall performance. 

\begin{figure*}[h!] % [h] 表示图片将尽可能放置在当前位置（here）
    \centering % 图片居中
    \includegraphics[width=\textwidth]{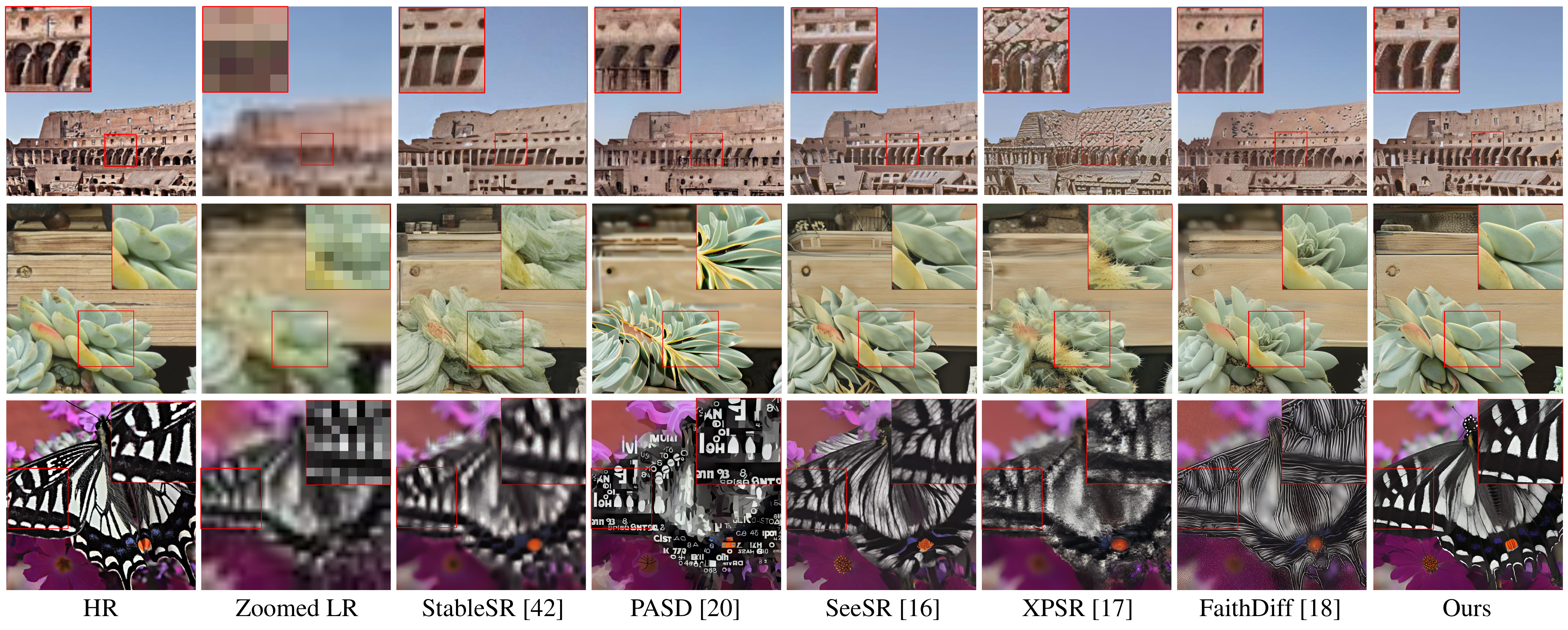} % 调整图片宽度为页面宽度的80%
    \caption{Qualitative comparisons under real-world degradations. SR scales from top to bottom: ×5.3, ×16, and ×24.} % 添加图片标题
    \label{fig:real} % 添加图片标签，便于引用
    % \vspace{2em}
\end{figure*}

\begin{table*}[htbp]
    \centering
    \renewcommand{\arraystretch}{0.9}
    \setlength{\tabcolsep}{4.5pt} % 调整列间距
    \caption{\sc Ablation on Scale-Controlled Mechanisms. The best and second-best results are highlighted in \textbf{\textcolor{red}{red}} and \textcolor{blue}{\underline{blue}}, respectively.}
    \begin{tabularx}{\textwidth}{p{25pt}|cc|ccccc|ccccc}
    \toprule
     & \textbf{Global} & \textbf{Local} & \multicolumn{5}{c|}{\textbf{DIV2K}} & \multicolumn{5}{c}{\textbf{DRealSR}} \\
     % \cmidrule(lr){4-8} \cmidrule(lr){9-13}
    \multirow{-2}{*}{\textbf{Scale $s$}} &  \textbf{Perception} & \textbf{Modulation} & PSNR$\uparrow$ & LPIPS$\downarrow$ & FID $\downarrow$ & MANIQA$\uparrow$ & CLIPIQA$\uparrow$ &  PSNR$\uparrow$ & LPIPS$\downarrow$ & FID$\downarrow$ & MANIQA$\uparrow$ & CLIPIQA$\uparrow$ \\
    \midrule
    
    & &  & \textbf{\textcolor{red}{22.62}} & \textbf{\textcolor{red}{0.2528}} & 120.26 & 0.5917 & \underline{\textcolor{blue}{0.7599}} & \textbf{\textcolor{red}{24.08}} & \textbf{\textcolor{red}{0.3109}} & \textbf{\textcolor{red}{158.55}} & 0.5822 & 0.6995 \\
    & & \checkmark & \underline{\textcolor{blue}{22.49}} & 0.2549 & \underline{\textcolor{blue}{115.66}} & 0.5855 & 0.7547 & 23.46 & 0.3177 & 163.63 & \underline{\textcolor{blue}{0.5899}} & 0.7041 \\
    & \checkmark &  & 21.85 & 0.2757 & 130.81 & \textbf{\textcolor{red}{0.6149}} & \textbf{\textcolor{red}{0.7788}} & 23.26 & 0.3197 & 168.99 & 0.5917 & \underline{\textcolor{blue}{0.7138}} \\
    \rowcolor{pink!20}
    \cellcolor{white} \multirow{-5}{*}{\parbox{12pt}{\centering ×5.3}} & \checkmark & \checkmark & 22.29 & \underline{\textcolor{blue}{0.2538}} & \textbf{\textcolor{red}{115.15}} & \underline{\textcolor{blue}{0.6021}} & 0.7548 & \underline{\textcolor{blue}{23.73}} & \underline{\textcolor{blue}{0.3127}} & \underline{\textcolor{blue}{161.83}} & \textbf{\textcolor{red}{0.5924}} & \textbf{\textcolor{red}{0.7147}} \\
    \midrule

    & &  & \textbf{\textcolor{red}{21.30}} & \underline{\textcolor{blue}{0.4212}} & 178.80 & 0.5093 & 0.7113 & \textbf{\textcolor{red}{23.36}} & 0.4453 & 201.91 & 0.5091 & 0.6774 \\
    & & \checkmark & \underline{\textcolor{blue}{20.46}} & 0.4340 & 171.19 & \underline{\textcolor{blue}{0.5637}} & \underline{\textcolor{blue}{0.7526}} & 22.39 & \textbf{\textcolor{red}{0.4529}} & \underline{\textcolor{blue}{198.26}} & \underline{\textcolor{blue}{0.5478}} & \underline{\textcolor{blue}{0.6998}} \\
    & \checkmark & & 19.71 & \textbf{\textcolor{red}{0.4196}} & \underline{\textcolor{blue}{165.72}} & 0.5527  & 0.7484 & 21.86 & 0.4648 & 204.61 & 0.5373 & 0.6840 \\
    \rowcolor{pink!20}
    \cellcolor{white} \multirow{-5}{*}{\parbox{12pt}{\centering ×16}} & \checkmark & \checkmark & 19.91 & 0.4238 & \textbf{\textcolor{red}{165.57}} & \textbf{\textcolor{red}{0.5685}} & \textbf{\textcolor{red}{0.7634}} & \underline{\textcolor{blue}{22.40}} & \underline{\textcolor{blue}{0.4593}} & \textbf{\textcolor{red}{194.00}} & \textbf{\textcolor{red}{0.5565}} & \textbf{\textcolor{red}{0.7111}} \\
    \midrule

    & &   & \textbf{\textcolor{red}{19.94}} & 0.4501 & 164.99 & 0.4571 & 0.6931 & \textbf{\textcolor{red}{22.73}} & \textbf{\textcolor{red}{0.4515}} & 186.45 & 0.4704 & 0.6487\\
    & & \checkmark & 19.67 & 0.4445 & 154.91 & 0.5317 & 0.7376 &  22.22 & 0.4633 & \underline{\textcolor{blue}{185.24}} & 0.5409 & 0.7093 \\
    & \checkmark & & 19.40 & \textbf{\textcolor{red}{0.4363}} & \underline{\textcolor{blue}{153.97}} & \underline{\textcolor{blue}{0.5515}} & \underline{\textcolor{blue}{0.7584}} & 22.08 & \underline{\textcolor{blue}{0.4552}} & 191.72 & \underline{\textcolor{blue}{0.5420}} & \underline{\textcolor{blue}{0.7169}} \\
    \rowcolor{pink!20}
    \cellcolor{white} \multirow{-5}{*}{\parbox{12pt}{\centering ×20.8}} & \checkmark & \checkmark & \underline{\textcolor{blue}{19.70}} & \underline{\textcolor{blue}{0.4402}} & \textbf{\textcolor{red}{153.32}} & \textbf{\textcolor{red}{0.5567}} & \textbf{\textcolor{red}{0.7628}} & \underline{\textcolor{blue}{22.62}} & 0.4646 & \textbf{\textcolor{red}{182.30}} & \textbf{\textcolor{red}{0.5422}} & \textbf{\textcolor{red}{0.7279}} \\
    \midrule

    & &  & \textbf{\textcolor{red}{19.74}} & 0.4789 & 164.31 & 0.4460 & 0.6797 & \underline{\textcolor{blue}{22.77}} & \textbf{\textcolor{red}{0.4670}} & 196.58 & 0.4401 & 0.6148\\
    & & \checkmark & 19.43 & 0.4720 & \underline{\textcolor{blue}{162.40}} & 0.5180 & 0.7404 & 22.26 & 0.4898 & \underline{\textcolor{blue}{188.57}} & \textbf{\textcolor{red}{0.5236}} & 0.6987 \\
    & \checkmark & & 19.33 & \textbf{\textcolor{red}{0.4625}} & 164.46 & \underline{\textcolor{blue}{0.5334}} & \underline{\textcolor{blue}{0.7497}} & 22.22 & \underline{\textcolor{blue}{0.4735}} & 191.44 & \underline{\textcolor{blue}{0.5205}} & \underline{\textcolor{blue}{0.7138}} \\
    \rowcolor{pink!20}
    \cellcolor{white} \multirow{-5}{*}{\parbox{12pt}{\centering ×24}} & \checkmark & \checkmark & \underline{\textcolor{blue}{19.69}} & \underline{\textcolor{blue}{0.4674}} & \textbf{\textcolor{red}{161.02}} & \textbf{\textcolor{red}{0.5339}} & \textbf{\textcolor{red}{0.7644}} & \textbf{\textcolor{red}{22.90}} & 0.4747 & \textbf{\textcolor{red}{187.10}} & 0.5111 & \textbf{\textcolor{red}{0.7268}} \\
    
    \bottomrule
    \end{tabularx}
    \label{tab:abl_scale}
    % \vspace{-1em}
\end{table*}

\begin{table*}[htbp]
    \centering
    \renewcommand{\arraystretch}{0.9}
    \setlength{\tabcolsep}{4pt} % 调整列间距
    \caption{\sc Comparison of different Semantic Extractors. The best and second-best results are highlighted in \textbf{\textcolor{red}{red}} and \textcolor{blue}{\underline{blue}}, respectively.} 
    \begin{tabularx}{\textwidth}{c|cc|cc|ccccc|ccccc}
    \toprule
     & \multicolumn{2}{c|}{\textbf{Textual-level}} & \multicolumn{2}{c|}{\textbf{Feature-level}} & \multicolumn{5}{c|}{\textbf{DIV2K}} & \multicolumn{5}{c}{\textbf{DRealSR}} \\
    % \cmidrule(lr){2-3}  \cmidrule(lr){4-5} 
    % \cmidrule(lr){6-10} \cmidrule(lr){11-15}
    \multirow{-2}{*}{\textbf{Scale $s$}} & Tag & Caption & $F_{sam}$ & $F_{ram}$ & PSNR$\uparrow$ & LPIPS$\downarrow$ & FID$\downarrow$ & MANIQA$\uparrow$ & CLIPIQA$\uparrow$ &  PSNR$\uparrow$ & LPIPS$\downarrow$ & FID$\downarrow$ & MANIQA$\uparrow$ & CLIPIQA$\uparrow$ \\
    \midrule
    
   & & & & \checkmark & 21.42 & 0.2891 & 135.37 & \underline{\textcolor{blue}{0.6164}} & \textbf{\textcolor{red}{0.7748}} & 23.08 & 0.3422 & 187.05 & 0.5890 & 0.6980  \\
    &  & \checkmark & & & 22.17 & 0.2763 &  128.03 & 0.5978 & 0.7595 & \underline{\textcolor{blue}{23.40}} & 0.3456 & 173.98 & 0.5933 & \textbf{\textcolor{red}{0.7192}} \\
    & \checkmark & & & \checkmark & \textbf{\textcolor{red}{22.37}} & \textbf{\textcolor{red}{0.2463}} & 120.19 & 0.5927 & 0.7652 & 23.21 & \underline{\textcolor{blue}{0.3350}} & 179.10 & \textbf{\textcolor{red}{0.6063}} & 0.7091 \\
    &  & \checkmark & \checkmark & & 21.62 & 0.2694 & \underline{\textcolor{blue}{118.98}} & \textbf{\textcolor{red}{0.6325}} & \underline{\textcolor{blue}{0.7745}} & 23.34 & 0.3380 & \underline{\textcolor{blue}{162.93}} & \underline{\textcolor{blue}{0.6021}} & \underline{\textcolor{blue}{0.7181}}\\
    \rowcolor{pink!20}
    \cellcolor{white} \multirow{-5}{*}{\parbox{12pt}{\centering ×5.3}}&  & \checkmark &  & \checkmark & \underline{\textcolor{blue}{22.29}} & \underline{\textcolor{blue}{0.2538}} & \textbf{\textcolor{red}{115.15}} & 0.6021 & 0.7548 & \textbf{\textcolor{red}{23.73}} & \textbf{\textcolor{red}{0.3127}} & \textbf{\textcolor{red}{161.83}} & 0.5924 & 0.7147 \\
    \midrule
    
    & & & & \checkmark & 19.19 & 0.4414 & 191.11 & 0.5227 & 0.7476 & \underline{\textcolor{blue}{21.61}} & 0.4571 & 210.76 & 0.4884 & 0.6606 \\
    &  & \checkmark & & & 19.27 & 0.4654 & 179.87 & 0.5443 & 0.7407 & 21.44 & 0.4871 & 207.74 & \underline{\textcolor{blue}{0.5402}} & 0.6876 \\
    & \checkmark & & & \checkmark & \underline{\textcolor{blue}{19.66}} & \underline{\textcolor{blue}{0.4293}} & 174.42 & \underline{\textcolor{blue}{0.5631}} & \underline{\textcolor{blue}{0.7598}} & 21.17 & \textbf{\textcolor{red}{0.4587}} & 212.24 & 0.4901 & 0.6524\\
    &  & \checkmark & \checkmark & & 19.17 & 0.4401 & \underline{\textcolor{blue}{168.75}} & 0.5558 & 0.7530 & 21.73 & 0.4770 & \textbf{\textcolor{red}{187.31}} & 0.5253 & \underline{\textcolor{blue}{0.6882}} \\
    \rowcolor{pink!20}
    \cellcolor{white} \multirow{-5}{*}{\parbox{12pt}{\centering ×16}}&  & \checkmark &  & \checkmark & \textbf{\textcolor{red}{19.91}} & \textbf{\textcolor{red}{0.4238}} & \textbf{\textcolor{red}{165.57}} & \textbf{\textcolor{red}{0.5685}} & \textbf{\textcolor{red}{0.7634}} & \textbf{\textcolor{red}{22.40}} & \underline{\textcolor{blue}{0.4593}} & \underline{\textcolor{blue}{194.00}} & \textbf{\textcolor{red}{0.5565}} & \textbf{\textcolor{red}{0.7111}} \\
    \midrule

    & & & & \checkmark & 19.08 & \underline{\textcolor{blue}{0.4661}} & 183.83 & 0.5030 & 0.7478 & 21.89 & \textbf{\textcolor{red}{0.4636}} & 194.85 & 0.4954 & 0.6997 \\
    &  & \checkmark & & & \underline{\textcolor{blue}{19.46}} & 0.4995 & 173.66 & 0.5312 & 0.7422 & 22.33 & 0.5179 & 194.57 & 0.5381 & \textbf{\textcolor{red}{0.7289}} \\
    & \checkmark & & & \checkmark & 19.28 & 0.4793 &  170.54 & \underline{\textcolor{blue}{0.5381}} & \underline{\textcolor{blue}{0.7571}} & 21.55 & 0.4979 & \textbf{\textcolor{red}{185.79}} & \underline{\textcolor{blue}{0.5160}} & 0.6978 \\
    &  & \checkmark & \checkmark & & 19.23 & \textbf{\textcolor{red}{0.4618}} &  \underline{\textcolor{blue}{162.67}} & \textbf{\textcolor{red}{0.5451}} & 0.7549 & \underline{\textcolor{blue}{22.49}} & 0.4944 & 189.67 & \textbf{\textcolor{red}{0.5212}} & 0.7155 \\
    \rowcolor{pink!20}
    \cellcolor{white} \multirow{-5}{*}{\parbox{12pt}{\centering ×24}}&  & \checkmark &  & \checkmark & \textbf{\textcolor{red}{19.69}} & 0.4674 & \textbf{\textcolor{red}{161.02}} & 0.5339 & \textbf{\textcolor{red}{0.7644}} & \textbf{\textcolor{red}{22.90}} & \underline{\textcolor{blue}{0.4747}} & \underline{\textcolor{blue}{187.10}} & 0.5111 & \underline{\textcolor{blue}{0.7268}} \\
    
    \bottomrule
    \end{tabularx}
    \label{tab: abl_sem}
    % \vspace{-0.5em}
\end{table*}

\begin{table*}[ht]
\setlength{\tabcolsep}{6pt} % 调整列间距
\renewcommand{\arraystretch}{0.9}
\centering
\caption{\sc Ablation on Auxiliary Fidelity Components. The best and second-best results are highlighted in \textbf{\textcolor{red}{red}} and \textcolor{blue}{\underline{blue}}, respectively.}
\begin{tabularx}{\textwidth}{c|cc|ccccc|ccccc}
\toprule
 &  &  & \multicolumn{5}{c|}{\textbf{DIV2K}} & \multicolumn{5}{c}{\textbf{DRealSR}} \\
 % \cline{3-7} \cline{8-12}
\multirow{-2}{*}{\textbf{Scale $s$}} & \multirow{-2}{*}{\textbf{$L_{LQA}$}} & \multirow{-2}{*}{\textbf{$M_{up}$}}& PSNR$\uparrow$ & LPIPS$\downarrow$ & FID$\downarrow$ & MANIQA$\uparrow$ & CLIPIQA$\uparrow$ &  PSNR$\uparrow$ & LPIPS$\downarrow$ & FID$\downarrow$ & MANIQA$\uparrow$ & CLIPIQA$\uparrow$ \\
\midrule
& \checkmark &  & 21.82 & 0.2626 & 121.27 & \textbf{\textcolor{red}{0.6424}} & \textbf{\textcolor{red}{0.7683}} & 23.49 & 0.3258 & 167.86 & \textbf{\textcolor{red}{0.6058}} & 0.7053 \\
&  & \checkmark & \textbf{\textcolor{red}{22.50}} & \underline{\textcolor{blue}{0.2577}} & \underline{\textcolor{blue}{116.80}} & \underline{\textcolor{blue}{0.6110}} & \underline{\textcolor{blue}{0.7631}} & \textbf{\textcolor{red}{23.90}} & 0.3166 & 162.77 & \underline{\textcolor{blue}{0.5983}} & \underline{\textcolor{blue}{0.7093}} \\
\rowcolor{pink!20}
\cellcolor{white} \multirow{-3}{*}{\parbox{14pt}{\centering ×5.3}}& \checkmark & \checkmark & \underline{\textcolor{blue}{22.29}} & \textbf{\textcolor{red}{0.2538}} & \textbf{\textcolor{red}{115.15}} & 0.6021 & 0.7548 & \underline{\textcolor{blue}{23.73}} & \textbf{\textcolor{red}{0.3127}} & \textbf{\textcolor{red}{161.83}} & 0.5924 & \textbf{\textcolor{red}{0.7147}} \\
\midrule
 & \checkmark &  & 19.30 & \textbf{\textcolor{red}{0.4186}} & 168.83 & 0.5429 & 0.7340 & \underline{\textcolor{blue}{21.98}} & 0.4629 & \textbf{\textcolor{red}{189.14}} & 0.5304 & \underline{\textcolor{blue}{0.6922}} \\
 &  & \checkmark & \underline{\textcolor{blue}{19.69}} & 0.4320 & \underline{\textcolor{blue}{165.84}} & \underline{\textcolor{blue}{0.5489}} & \underline{\textcolor{blue}{0.7376}} & \underline{\textcolor{blue}{21.98}} & \underline{\textcolor{blue}{0.4614}} & \underline{\textcolor{blue}{192.53}} & \underline{\textcolor{blue}{0.5332}} & 0.6871\\
 \rowcolor{pink!20}
\cellcolor{white} \multirow{-3}{*}{\parbox{14pt}{\centering ×16}}& \checkmark & \checkmark & \textbf{\textcolor{red}{19.91}} & \underline{\textcolor{blue}{0.4238}} & \textbf{\textcolor{red}{165.57}} & \textbf{\textcolor{red}{0.5685}} & \textbf{\textcolor{red}{0.7634}} & \textbf{\textcolor{red}{22.40}} & \textbf{\textcolor{red}{0.4593}} & 194.00 & \textbf{\textcolor{red}{0.5565}} & \textbf{\textcolor{red}{0.7111}} \\
\midrule
 & \checkmark &  & 18.54  & 0.5143 & 173.04 & \textbf{\textcolor{red}{0.5527}} & 0.7331 & 21.78 & 0.5653 & 202.00 & \textbf{\textcolor{red}{0.5333}}& 0.6839 \\
 &  &  \checkmark & \underline{\textcolor{blue}{19.38}} & \textbf{\textcolor{red}{0.4663}} & \underline{\textcolor{blue}{161.48}} & 0.5275 & \underline{\textcolor{blue}{0.7409}} & \underline{\textcolor{blue}{22.19}} & \underline{\textcolor{blue}{0.4941}} & \underline{\textcolor{blue}{189.50}} & \underline{\textcolor{blue}{0.5244}} & \underline{\textcolor{blue}{0.7160}} \\
 \rowcolor{pink!20}
\cellcolor{white} \multirow{-3}{*}{\parbox{14pt}{\centering ×24}}& \checkmark & \checkmark & \textbf{\textcolor{red}{19.69}} & \underline{\textcolor{blue}{0.4674}} & \textbf{\textcolor{red}{161.02}} & \underline{\textcolor{blue}{0.5339}} & \textbf{\textcolor{red}{0.7644}} & \textbf{\textcolor{red}{22.90}} & \textbf{\textcolor{red}{0.4747}} & \textbf{\textcolor{red}{187.10}} & 0.5111 & \textbf{\textcolor{red}{0.7268}} \\
\bottomrule
\end{tabularx}
% \vspace{-1em}
\label{tab:abl_fie}
\end{table*}

\subsection{Quantitative Evaluation} 
We provide quantitative comparisons in Tables~\ref{tab: simple} and~\ref{tab: real-world}. Table~\ref{tab: simple} shows results on bicubic downsampling datasets, and Table~\ref{tab: real-world} presents results on real-world degradation benchmarks, including synthetic and real-captured datasets.

Although non-diffusion-based methods generally achieve higher PSNR values across all SR scales, as shown in Table~\ref{tab: simple}, their overall performance remains inferior to diffusion-based approaches when evaluated from perceptual similarity (LPIPS, DISTS), distribution similarity (FID), and no-reference image quality (MANIQA, MUSIQ, CLIPIQA). In addition, they cannot handle complex real-world degradations.

IDM is a pixel-domain diffusion that equips INR to achieve ASSR, which is also a method designed for ideal downsampling degradation without real-world generation ability. Although it demonstrates better non-reference metrics (MANIQA, MUSIQ, CLIPIQA) than non-diffusion-based methods, it still significantly lags behind other diffusion-based Real-ISR methods that leverage pre-trained diffusion prior. StableSR and PASD show clear improvements. However, as the SR scale increases, their results show inferior results than recent methods such as SeeSR, XPSR, FaithDiff, and our OmniScaleSR, with larger discrepancies observed in both fidelity and realism metrics.

As shown in Tables~\ref{tab: simple} and~\ref{tab: real-world}, XPSR and FaithDiff are the two most competitive methods, but both struggle to perform well on both fidelity and realism metrics. While XPSR performs well on MANIQA and CLIPIQA, sometimes even surpassing our method, these indicators are prone to being inflated due to excessive and cluttered details \cite{wu2024seesr}, as can be seen from the qualitative results of XPSR in Fig. \ref{fig:clean} and Fig. \ref{fig:real}. This is also reflected in XPSR’s qualitative results in Fig.~\ref{fig:clean} and Fig.~\ref{fig:real}. This is further corroborated by XPSR’s markedly lower LPIPS, FID, and DISTS scores compared with SeeSR, FaithDiff, and our method, indicating inferior fidelity. On the other hand, FaithDiff demonstrates better fidelity at low SR scales in terms of LPIPS and DISTS, but exhibits reduced realism at higher SR scales (above ×16), yielding substantially lower MANIQA and CLIPIQA scores compared with our method.

In contrast, our method, OmniScaleSR, consistently achieves both high fidelity and realism across diverse SR scales, degradation models, and datasets. It ranks first or second in nearly all metrics, demonstrating superior overall performance. Notably, for ultra-high SR scales (such as ×24 on RealSR), where XPSR's fidelity and FaithDiff's realism degrade significantly, our approach maintains both qualities. It surpasses XPSR by 12.9\% in LPIPS and FaithDiff by 16.6\% in CLIPIQA, highlighting our advantage at extreme SR scales.

\begin{figure*}[htbp] % [h] 表示图片将尽可能放置在当前位置（here）
    \centering % 图片居中
    \includegraphics[width=\textwidth]{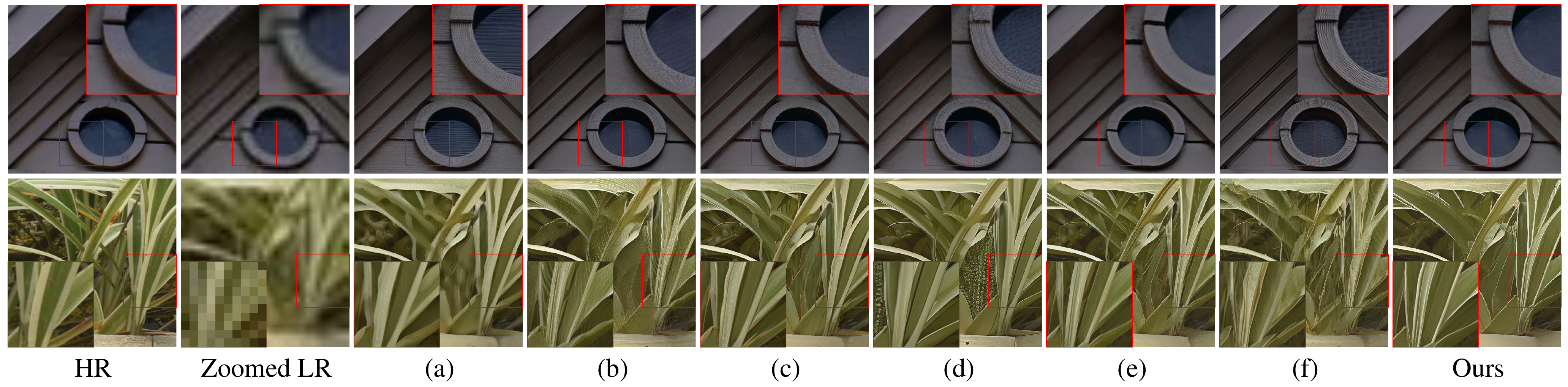} % 调整图片宽度为页面宽度的80%
    \caption{\textbf{Ablation study on key components of our method}: (a) without all scale-controlled mechanisms, (b) without local scale modulation mechanism, (c) without global scale injection mechanism, (d) without SePR Attention, (e) baseline using tags to replace the image captions as text prompt, (f) without the pre-trained ×4 upsampler.} % 添加图片标题
    \label{fig:ablation} % 添加图片标签，便于引用
    % \vspace{-1em}
\end{figure*}

\begin{table}[tbp]
\setlength{\tabcolsep}{4pt} % 调整列间距
\renewcommand{\arraystretch}{0.9}
\centering
\caption{\sc Ablation on Sampling Strategies. The best results are highlighted in \textbf{Bold}.}
\begin{tabular}{c|ccccc}
\toprule
 scheduler & PSNR$\uparrow$ & LPIPS$\downarrow$ & FID$\downarrow$ & MANIQA$\uparrow$ & CLIPIQA$\uparrow$ \\
\midrule
 DPM-Solver & 18.34 & 0.4578 & 166.04 & 0.5086 & 0.7319 \\
 DDIM & 18.67 & 0.4591 & \textbf{164.67} & 0.5616 & 0.7598 \\
 \rowcolor{gray!10}
 DDPM (Ours) & \textbf{19.91} & \textbf{0.4238} & 165.57 & \textbf{0.5685} & \textbf{0.7634} \\
\bottomrule
\end{tabular}
\label{tab:sample}
% \vspace{-2em}
\end{table}

\subsection{Qualitative Comparison}
We provide qualitative comparison results on bicubic degradation and real-world degradation in Fig.\ref{fig:clean} and Fig.\ref{fig:real}, respectively. As shown in Fig. \ref{fig:clean}, non-diffusion-based ASSR methods generally preserve the overall scene structure and avoid introducing hallucinated content. However, their outputs tend to be overly smooth and lack realistic, fine-grained textures, resulting in blurry and visually unsatisfactory results. Although IDM is diffusion-based, it does not exploit generative diffusion prior and lacks latent-space modeling, leading to limited realism and poor performance as previously discussed.

StableSR and PASD yield more realistic results. However, as the SR scale increases, both methods suffer from significant distortions such as structural deformation, hallucinated textures, and aliasing artifacts, as shown in Fig. \ref{fig:clean} and Fig. \ref{fig:real}. SeeSR demonstrates improved structural and textural fidelity, and show better robustness to higher SR scales. However, it still suffers from significant distortions and artifacts at extreme SR scales such as ×24.

XPSR and FaithDiff demonstrate improved overall performance than StableSR, PASD and SeeSR. But they still cannot handle various SR scales, particularly extreme-large SR scales. XPSR tends to hallucinate inconsistent semantics and structures, while FaithDiff often produce blurry results.

In contrast, compared with them, our OmniScaleSR consistently produces high-quality images with faithful reconstructions and realistic details across all SR scales and demonstrates significantly better performance at extreme magnifications like ×24, achieving both significantly better fidelity and realism.

\subsection{Ablation Studies} 

In this section, we conduct ablation studies to systematically validate the contributions of our proposed components.  
All experiments are conducted on both the DIV2K and DRealSR datasets, representing simple downsampling and real-world degradation scenarios, respectively. Ablation experiments on DIV2K use a representative subset of 100 images, which is sufficient to reveal the effects of each component.

\subsubsection{Effects of Scale-Controlled Mechanisms}
We provide ablation study results on the two proposed scale-controlled mechanisms in Table \ref{tab:abl_scale}. Enabling either of them individually leads to noticeable improvements in both fidelity (PSNR, LPIPS) and perceptual quality (FID, MANIQA, CLIPIQA) compared to the baseline, particularly at higher SR scales. For instance, at ×24 on DIV2K, incorporating either of them reduces LPIPS by 0.01 and improves CLIPIQA by 0.07, demonstrating their independent contributions. 
Combining both of them yields the best or second-best performance across most metrics. Although using the scale-controlled mechanisms results in a slight sacrifice in PSNR and LPIPS at ×5.3, the overall improvement is significant. Fig. \ref{fig:ablation} shows qualitative results from ×5.3 to ×16. It is evident that the removal of scale-controlled mechanisms results in noticeable blurriness and weird texture artifacts, and the best visual quality is achieved when both mechanisms are combined.

\begin{table*}[htbp]
\setlength{\tabcolsep}{14pt} % 调整列间距
\renewcommand{\arraystretch}{0.9}
\centering
\caption{\sc Complexity comparison among different methods. All methods are evaluated using 4× SR on input images of size 128 × 128, and the inference time is measured on an A100 GPU. 'S' denotes the number of diffusion inference steps.}
\begin{tabular}{c|cccccccc}
\toprule
& IDM & StableSR & PASD & SeeSR & XPSR & FaithDiff & Ours & Ours-S20 \\
\midrule
Inference Step & 200 & 200 & 20 & 50 & 20 & 20 & 50 & 20 \\
Inference Time (s) & 14.93 & 11.89 & 4.84 & 5.82 & 11.43 & 8.53 & 6.96 & 4.98 \\
% MACs (G) &154606 & 79940 & 29125 & 65857 & 28046 & 65947 \\
\# Trainable Param (M) &  116.6 & 150.0 & 625.0 & 749.9 & 868.5 & 2612.2 &790.1 & 790.1 \\
\# Total Param (B) &  0.12 & 1.56 & 2.31 & 2.52 & 15.95 & 15.70 & 5.25 & 5.25 \\
\bottomrule
\end{tabular}
% \vspace{-1em}
\label{tab:complexity}
\end{table*}

\subsubsection{Impact of Different Semantic Extractors}
We apply dual-level semantics to improve the fidelity and overall performance. Specifically, we use BLIP-2 \cite{li2023blip} to extract text prompts and a fine-tuned RAM \cite{zhang2024recognize} to extract feature prompts. Here we investigate the impact of different semantic extractors and the results are shown in Table~\ref{tab: abl_sem}. We compare our method with various variants, including only using text prompts, only using feature prompts, and using both but different extractors. 'Tag' refers to utilizing fine-tuned RAM to extract tags as text prompts, and '$F_{sam}$' indicates the use of the SAM-Tiny \cite{ravi2024sam2} image encoder for extracting semantic features. Results show that removing either semantic guidance results in a moderate performance drop, highlighting their complementary roles. Combining both prompts consistently yields the best results, with our adopted setup achieving the best performance. Visual comparisons in Fig.~\ref{fig:ablation} (d), (e), and 'ours' confirm this.

\subsubsection{Ablation on Auxiliary Fidelity Components} Table \ref{tab:abl_fie} provides the quantitative ablation results on LQA loss and the pre-trained ×4 SR model. The LQA loss consistently improves performance across all SR scales, enhancing both fidelity and realism. The pre-trained ×4 SR model $M_{up}$, with only 232K parameters, plays a crucial role in enhancing the fidelity of high-magnification SR. On the DRealSR dataset, it reduces LPIPS by 0.09 at the extreme SR scale of ×24, clearly demonstrating the importance of providing the diffusion fidelity branch with a high-quality initial input. Fig. \ref{fig:ablation} (f) shows the qualitative results without $M_{up}$, where fidelity decreases compared to the full model.

\subsubsection{Ablation on Sampling Strategies} 
Table \ref{tab:sample} presents the results of three different sampling strategies: DPM-Solver \cite{lu2022dpm}, DDIM \cite{song2021denoising}, and DDPM \cite{ho2020denoising} on the DIV2K dataset at a ×16 SR scale. Since the trends across other SR scales are consistent, we show the results for the ×16 SR scale as a representative case. As shown, our adopted DDPM achieves the best performance.

\begin{figure}[tb] % [h] 表示图片将尽可能放置在当前位置
    \centering % 图片居中
    \includegraphics[width=0.5\textwidth] {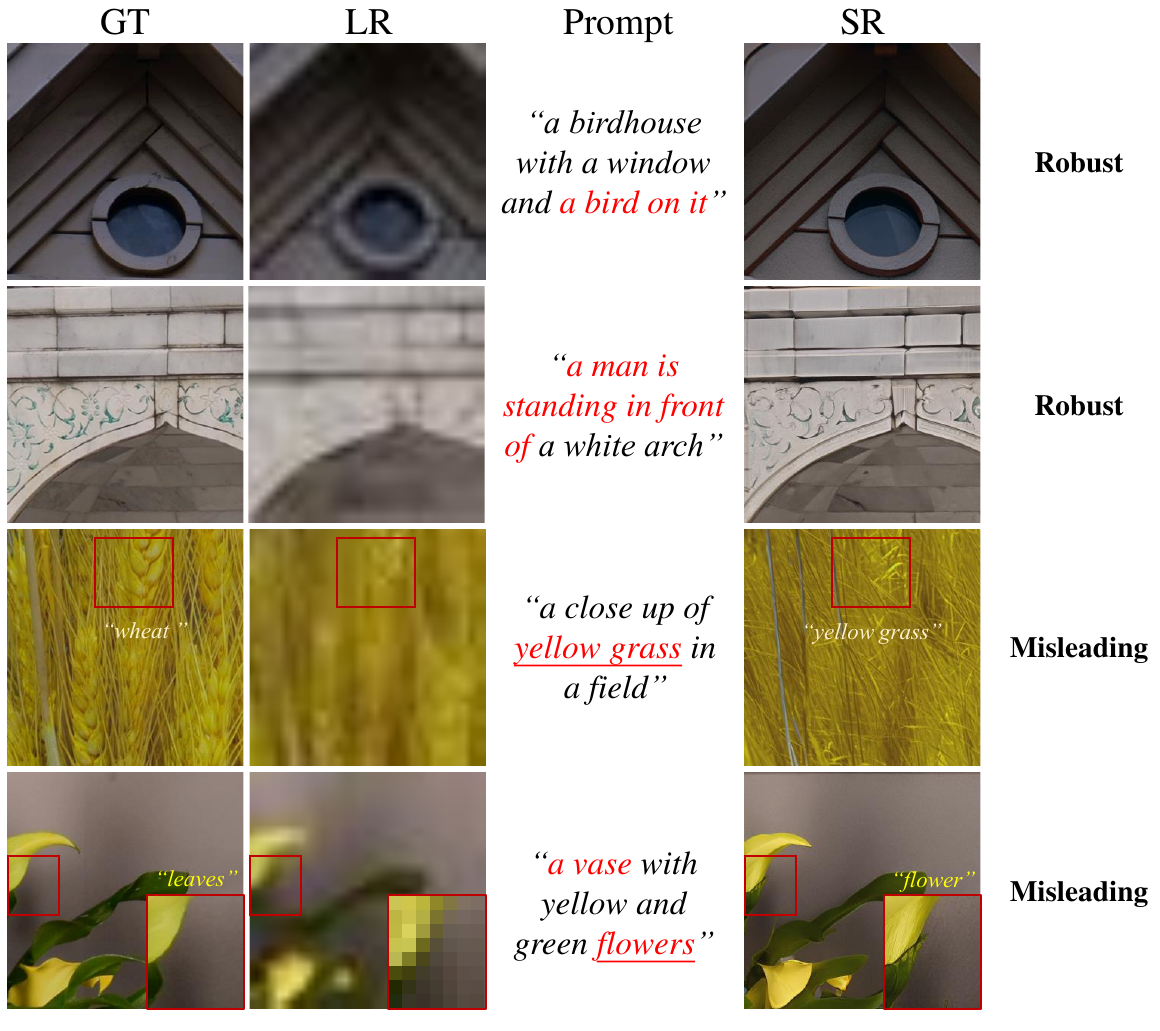} % 调整图片宽度为页面宽度的80%
    \caption{\textbf{Potential Risk Analysis under Inaccurate Prompts:} Inaccurate descriptions are highlighted in \textit{\textcolor{red}{red}}. Our method generally maintains high fidelity despite inaccurate prompts (top two rows). However, in some challenging cases where the incorrect prompt is similar in both textual semantics and visual structures, differing only in details, it may mislead the reconstruction process (bottom two rows), and we highlight these misleading words in \textit{\textcolor{red}{\underline{underline}}}.
    } % 添加图片标题
    \label{fig:promptbias} % 添加图片标签，便于引用
    \vspace{-1em}
\end{figure}

\subsubsection{Potential Risk Analysis Under Inaccurate Prompts} 
We employ BLIP-2 generated prompts as semantic guidance to improve the overall performance. However, inaccurate captions may introduce semantic bias and mislead reconstruction. Therefore, we further analyze the model’s robustness to inaccurate prompts. As illustrated in Fig. \ref{fig:promptbias}, in most cases, our method maintains high semantic, structural, and textural fidelity despite prompt inaccuracies (top two rows), benefiting from the multi-domain fidelity enhancement designs. However, in some challenging cases where the incorrect prompt is similar in both textual semantics and visual structure, differing only in details, it may mislead the reconstruction process. For example, in the third row, the prompt “yellow grass” leads to the misinterpretation of wheat as grass. Such misguidance in challenging cases remains a limitation of our current approach and a common issue among prompt-guided methods.

\subsection{Complexity Analysis}
We provide a complexity comparison with competing diffusion-based methods in Table \ref{tab:complexity}. The inference time is measured for the ×4 SR task with 128×128 LR images using one NVIDIA A100 GPU, including the time for text prompt extraction. “Ours-S20” refers to a variant of our method configured with 20 denoising steps, provided as a reference to enable fair complexity comparison under equal step settings. IDM and StableSR have longer inference times due to their use of 200 denoising steps. PASD, SeeSR, XPSR, and our method have similar numbers of trainable parameters, as they all utilize the pre-trained SD-1.5/SD-2 model, with differences primarily in the control components. The trainable parameters refer to the additional design introduced by each method. The larger total parameter count of our method primarily stems from the use of BLIP-2 (2.7B) for text prompt extraction. FaithDiff and XPSR have the largest total parameter counts, as they use LLaVA \cite{liu2023visual} to extract prompts. FaithDiff uses the 13B model, and XPSR uses two 7B models. The use of LLaVA also constitutes a major portion of their inference time. Moreover, FaithDiff employs a more powerful pre-trained model SDXL \cite{podell2024sdxl}, and unfreezes all parameters during the second-stage training, resulting in the largest number of trainable parameters. In summary, our method has a larger parameter count and inference time compared to PASD and SeeSR, but it offers significant advantages by saving both time and parameters compared to XPSR and FaithDiff.

\section{Conclusion and Limitations}
In this paper, we introduce OmniScaleSR, a novel diffusion-based Real-ASSR method, unleashing scale-controlled diffusion prior to achieving both high-fidelity and high-realism ASSR. We leverage the pre-trained diffusion prior for implicit generative SR scale adaptation, and introduce explicit diffusion-native SR scale control mechanisms to actively regulating the latent diffusion process with the target SR scale. Specifically, we propose a global scale injection mechanism and a local scale modulation mechanism. Additionally, we introduce multi-domain fidelity enhancement designs to reinforce structural and textural consistency. Extensive experimental results across both bicubic degradation benchmarks and real-world data demonstrate that OmniScaleSR consistently outperforms existing methods across various SR scales.

The main limitation of our method lies in its relatively long inference time. Incorporating recent advances in accelerating diffusion-based models, such as one-step acceleration \cite{wu2024onestep,dong2025tsd,sun2025pixel,yue2025arbitrary} and quantization strategies \cite{zhu2025passionsr,chen2025faithdiff}, is a promising direction to further improve efficiency. Secondly, similar to other diffusion-based restoration methods, our approach does not guarantee strict pixel-level consistency in terms of PSNR. Therefore, it may not be ideal for scenarios where extremely high pixel-level fidelity reconstruction is required. In addition, inaccurate prompts may lead to semantic bias, which is also a common limitation of prompt-guided methods. Our approach remains robust in most cases, but in some challenging cases where the incorrect prompt is similar in both the textual semantics and visual structure but differs in fine details, the incorrect prompt may mislead the reconstruction. Incorporating adaptive prompt reliability estimation or corrective modulation remains a promising direction for future research.

% \section*{Acknowledgments}
% This should be a simple paragraph before the References to thank those individuals and institutions who have supported your work on this article.

\bibliographystyle{unsrt}
\bibliography{reference}

\end{document}